\documentclass[letterpaper]{article} 
\usepackage{aaai2026}  
\usepackage{times}  
\usepackage{helvet}  
\usepackage{courier}  
\usepackage[hyphens]{url}  
\usepackage{graphicx} 
\usepackage{natbib}  
\usepackage{caption} 
\usepackage{algorithm}
\usepackage{algorithmic}

\usepackage{newfloat}
\usepackage{listings}
\DeclareCaptionStyle{ruled}{labelfont=normalfont,labelsep=colon,strut=off} 
\floatstyle{ruled}
\newfloat{listing}{tb}{lst}{}
\floatname{listing}{Listing}
\usepackage{subcaption}
\usepackage{booktabs}
\usepackage{multirow}
\usepackage{makecell}

\title{The Curious Case of Analogies: Investigating Analogical Reasoning\\in Large Language Models}
\author{
    Taewhoo Lee\textsuperscript{\rm 1,\rm 2},
    Minju Song\textsuperscript{\rm 1},
    Chanwoong Yoon\textsuperscript{\rm 1},
    Jungwoo Park\textsuperscript{\rm 1,\rm 2},
    Jaewoo Kang\textsuperscript{\rm 1,\rm 2}\thanks{Corresponding author.}
}
\affiliations{
    \textsuperscript{\rm 1}Korea University\\
    \textsuperscript{\rm 2}AIGEN Sciences\\

    \{taewhoo, minjusong, cwyoon99, jungwoo-park, kangj\}@korea.ac.kr
}

\begin{document}

\maketitle 

\begin{abstract}

Analogical reasoning is at the core of human cognition, serving as an important foundation for a variety of intellectual activities. 
While prior work has shown that LLMs can represent task patterns and surface-level concepts, it remains unclear whether these models can encode high-level relational concepts and apply them to novel situations through structured comparisons.
In this work, we explore this fundamental aspect using proportional and story analogies, and identify three key findings. 
First, LLMs effectively encode the underlying relationships between analogous entities; both attributive and relational information propagate through mid-upper layers in correct cases, whereas reasoning failures reflect missing relational information within these layers.
Second, unlike humans, LLMs often struggle not only when relational information is missing, but also when attempting to apply it to new entities. In such cases, strategically patching hidden representations at critical token positions can facilitate information transfer to a certain extent.
Lastly, successful analogical reasoning in LLMs is marked by strong structural alignment between analogous situations, whereas failures often reflect degraded or misplaced alignment.
Overall, our findings reveal that LLMs exhibit emerging but limited capabilities in encoding and applying high-level relational concepts, highlighting both parallels and gaps with human cognition.
\end{abstract}
\begin{links}
    \link{Code}{https://github.com/dmis-lab/analogical-reasoning}
\end{links}
\section{Introduction}
\label{section:introduction}

Analogical reasoning is a fundamental aspect of human cognition, enabling humans to navigate unfamiliar situations by drawing parallels to familiar concepts~\cite{hofstadter2001epilogue, holyoak2001place, hofstadter2013surfaces}. This ability serves as the foundation for a wide range of cognitive functions, including knowledge adaptation~\cite{doi:10.1080/713755671}, problem solving, and creative thinking~\cite{jbp:/content/journals/10.1075/pc.4.2.12gen}. 
Among various types of analogies, proportional analogies are widely used to assess one's ability to extract semantic relationships and apply them to new contexts~\cite{brown1989two}. 
For example, given the query ``\textit{Persuasion is to Jane Austen as 1984 is to}", one would first focus on the first pair of entities (\textit{Persuasion}, \textit{Jane Austen}) to identify the semantic relationship (``\textit{author of}"), and apply it to the third entity (\textit{1984}) to obtain the correct answer (``\textit{George Orwell}"). 
Extending this rudimentary setting, the ability to draw parallels between situations can be evaluated using story analogies. For instance, despite different surface details between ``\textit{missing} a train but \textit{encountering} a dear friend" and ``\textit{getting injured} but \textit{coming back stronger}", we find corresponding elements binded under the same theme: that every cloud has a silver lining.

\begin{figure}[t!]
\centering
\includegraphics[width=\columnwidth]{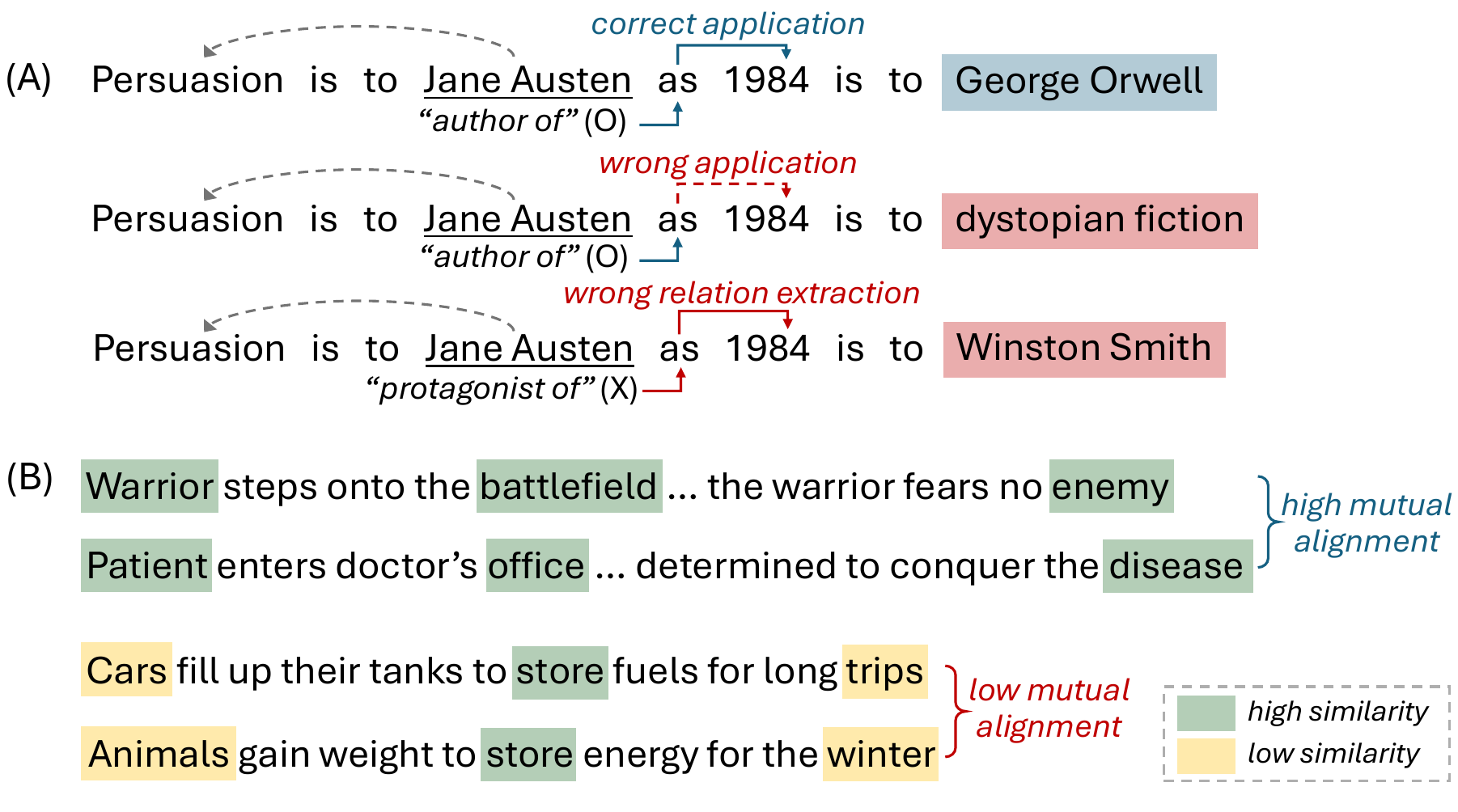}
\caption{
An overview of the mechanism behind analogical reasoning in LLMs. (A) LLMs effectively encode relational information and apply it during correct analogical reasoning, but applying the relation often remains as much a bottleneck as encoding it. (B) Identifying analogous situations is strongly associated with structural alignment, which we quantify using the Mutual Alignment Score (MAS).
}
\label{fig:motivation}
\end{figure}

Meanwhile, the advent of large language models (LLMs) and their remarkable performance on various tasks have spurred interest in the research community. 
Trained on massive text corpora with billions of parameters, modern LLMs have shifted the paradigm of problem-solving from task-specific fine-tuning to leveraging instructions and examples in the input prompt~\cite{brown2020language}. 
This emergent ability has motivated researchers to explore LLMs for complex reasoning tasks in diverse domains~\cite{kojima2022large, yao2023react, imani2023mathprompter, NEURIPS2023_631bb943}. Recently, there has been growing interest in the analogical reasoning capabilities of LLMs, focusing on evaluating~\cite{wijesiriwardene2023analogical, webb2023emergent} or advancing~\cite{wijesiriwardene2024exploring} these capabilities. 
However, the inner mechanisms behind LLMs and their ability to perform analogical reasoning remains unexplored. How do models extract relationships and apply them to predict the correct answer? Moreover, how do they draw parallels between semantically disparate, yet analogous context?

In this work, we take a closer look at how modern LLMs perform analogical reasoning. We first examine how information that bridges different entities is extracted and applied using proportional analogies. To understand this information flow, we analyze where critical signals are processed within the input. By blocking the final token from attending to different token positions, we find that mid-upper layers within the second and third entities (e.g., “Jane Austen” and “1984” in Figure~\ref{fig:motivation}) carry essential information; disrupting these positions leads to noticeable drops in performance. Further analysis shows that these positions encode both attributive and relational information, with relational content showing a significant gap between correct and incorrect cases. This suggests that, much like humans, models can not only represent individual entities but also abstract the underlying relation that connects them, highlighting relational reasoning as a central mechanism in analogical understanding.

Extending this analysis, we find that applying relational information poses an additional challenge beyond merely identifying it. Replacing the initial entity pairs in incorrect cases with those from correct ones changes model behavior in up to 38.4\% of cases, suggesting that models still struggle to transfer relational structure in the remaining cases. Building on earlier insights about the role of linking positions (e.g., “as”), we conduct patching interventions to facilitate information flow between entity pairs. These adjustments lead to successful answer revisions in up to 38.1\% of the remaining cases, highlighting that failures in analogical reasoning stem not only from representational gaps but also from limitations in relational application.

To deepen our understanding of how models perform analogical reasoning, we turn to the question of structural alignment, i.e., how models identify and map high-level relational parallels between seemingly unrelated concepts. Using story analogies, we reveal that analogical structure becomes increasingly linearly separable in the middle layers, and that successful reasoning is associated with stronger token-level alignment between source and target stories, despite minimal lexical overlap. These findings suggest that, beyond encoding entity-level information, LLMs develop abstract relational representations and perform alignment operations that mirror core aspects of human analogical reasoning.

In summary, our contributions include:
\begin{itemize}
    \item We investigate the internal mechanisms of LLMs in analogical reasoning, focusing on how models succeed (or fail) to extract and apply relational information.
    \item We analyze how structural alignment emerges in model representations, associating it with deeper token-level alignment between analogous situations.
    \item We contextualize model behavior by comparing it with human cognition, highlighting both parallels in relational abstraction and limitations in alignment and application.
\end{itemize}

\section{Preliminaries}
\label{section:preliminaries}

In this section, we provide an overview of prior research on analogical reasoning (see Section~\ref{subsec:analogical_reasoning_preliminaries}). We then discuss studies in mechanistic interpretability, focusing on methods used in our research (see Section~\ref{subsec:mech_interp_preliminaries}). Lastly, we clarify key terminologies used throughout the paper (see Section~\ref{subsec:terminology_preliminaries}).

\subsection{Analogical Reasoning}
\label{subsec:analogical_reasoning_preliminaries}

Analogical reasoning is a cognitive process that requires identifying relational similarities to understand new situations, form abstract concepts, and draw on past experiences to tackle novel problems~\cite{boteanu2015solving}.
Analogies can take several forms, including word analogies~\cite{gladkova-etal-2016-analogy, yuan-etal-2024-analogykb}, proportional analogies~\cite{mikolov-etal-2013-linguistic}, story analogies~\cite{jiayang-etal-2023-storyanalogy}, and long-text analogies~\cite{sultan-shahaf-2022-life}.
In this work, we focus on two types that best represent the cognitive requirements of analogical reasoning: proportional analogies, which require extracting and applying semantic relationships in the form ``\textit{A is to B as C is to D}"; and story analogies, which demand structural alignment between semantically distinct narratives or situations.

In the field of natural language processing (NLP), analogical reasoning has been explored through both benchmark construction~\cite{ye-etal-2024-analobench, jiayang-etal-2023-storyanalogy} and behavioral evaluation~\cite{webb2023emergent}. Others propose prompting strategies to leverage analogical capabilities more effectively, such as self-generated exemplars~\cite{yasunaga2024large} or knowledge-enhanced prompts~\cite{wijesiriwardene2024exploring}.
In a related line of work, several studies have examined how LLMs encode abstract task-level information when presented with in-context examples~\cite{hendel-etal-2023-context, todd2024function, opielka2025analogical}. These works identify task or function vectors, i.e., compact representations that reflect the operation demonstrated in ICL settings. While these studies provide evidence that models can internally represent conceptual relations, they are primarily limited to simple tasks (e.g., color matching, antonyms) and focus on detecting the presence of these vectors rather than analyzing what they represent and how they are used in more complex reasoning scenarios. In contrast, our work directly targets analogical reasoning behavior, offering a comprehensive view on how models extract, apply, and structurally align relational information.

\subsection{Mechanistic Interpretability}
\label{subsec:mech_interp_preliminaries}

Understanding the internal mechanisms of LLMs has been a central focus of recent research ~\cite{bereska2024mechanistic}. 
Among the various techniques developed to analyze intermediate activations and their causal roles in model behavior, two broad categories are particularly relevant to our work: representational analysis~\cite{logit_lens, tuned_lens, future_lens}, which investigates what types of information are encoded in hidden states; and intervention-based methods~\cite{causal_mediation, towards_activation_patching, paragraph_activation}, which manipulate internal activations to examine their functional impact on model outputs. Our study builds on both paradigms to probe the internal computations that support analogical reasoning. Below, we introduce key methods employed in our experiments:

\textbf{(1) Attention Knockout}~\cite{wang2023interpretability, geva-etal-2023-dissecting}: This method involves selectively disabling attention heads to examine their contribution to predicting outputs. By removing specific attention pathways, we can assess whether specific tokens are responsible for prediction correct outputs and identify which components are crucial for resolving relational information.

\textbf{(2) Linear Probing}~\cite{alain2018understandingintermediatelayersusing, belinkov-2022-probing}: This technique assesses whether specific types of information are linearly separable within a model’s hidden representations. Given labeled examples, we extract activation vectors from a particular layer and train a linear classifier to predict the labels. High probe accuracy suggests that the relevant information is explicitly encoded in the representation space at that layer.

\textbf{(3) Patchscopes}~\cite{10.5555/3692070.3692690}: A recent extension of activation patching that leverages the model's generative capabilities to interpret what information is encoded in its hidden representations. 
Specifically, a source prompt is first passed through the model, and the hidden representation of the token we wish to inspect is recorded. Next, the same model processes a \textit{target prompt}, which is used to induce natural language descriptions regarding the representation. For example, when using the target prompt constructed by~\citet{10.5555/3692070.3692690}: ``\texttt{Syria: Country in the Middle East, Leonardo DiCaprio: American actor, Samsung: South Korean multinational major appliance and consumer electronics corporation, x}", the representation in ``\texttt{x}" is replaced with the previously recorded representation, resulting in the description of that representation.
Throughout this work, we systematically construct diverse target prompts suitable for extracting different types of information. 

\subsection{Terminology}
\label{subsec:terminology_preliminaries}
In our experiments, proportional analogies follow the structure of ``$e_1$ is to $e_2$ as $e_3$ is to $e_4$". We refer to ``as" as the \textit{link}, and the underlying connection that groups entities together (e.g., ``author of" in ``\textit{Persuasion is to Jane Austen}") as the \textit{relation}. Story analogies include a source story, a target story (analogous), and a distractor story (lexically similar). For both settings, we refer to the final position of the input as the \textit{resolution token}.
\section{Experimental Setup}
\label{section:exp_setup}
\begin{figure*}[t!]
\centering
\begin{subfigure}[b]{0.49\textwidth}
    \centering
    \resizebox{0.8\textwidth}{!}{ 
        \includegraphics{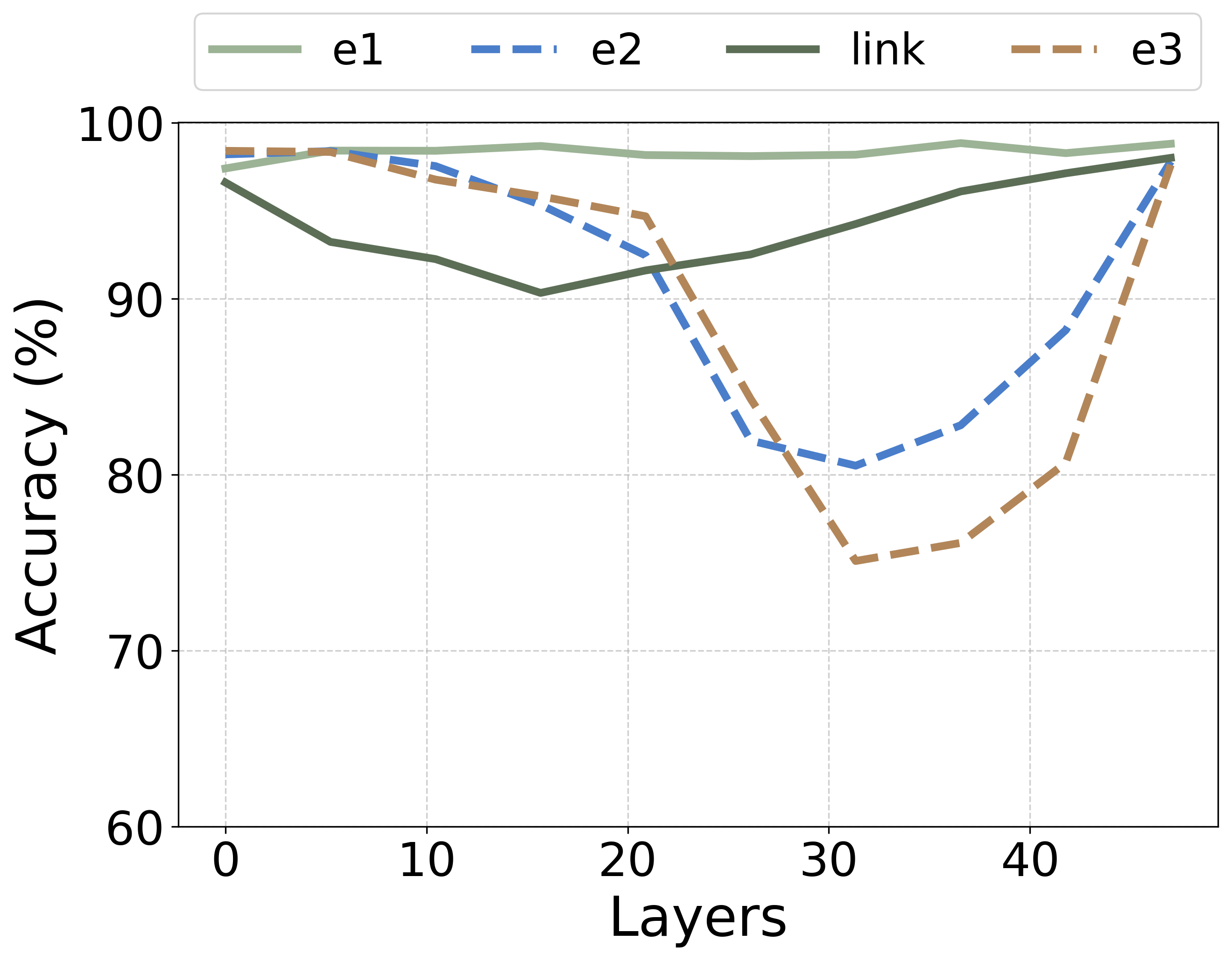}
    }
    \caption{Correct case} 
    \label{fig:curved_4_2_a}
\end{subfigure}
\hfill
\begin{subfigure}[b]{0.49\textwidth}
    \centering
    \resizebox{0.8\textwidth}{!}{ 
        \includegraphics{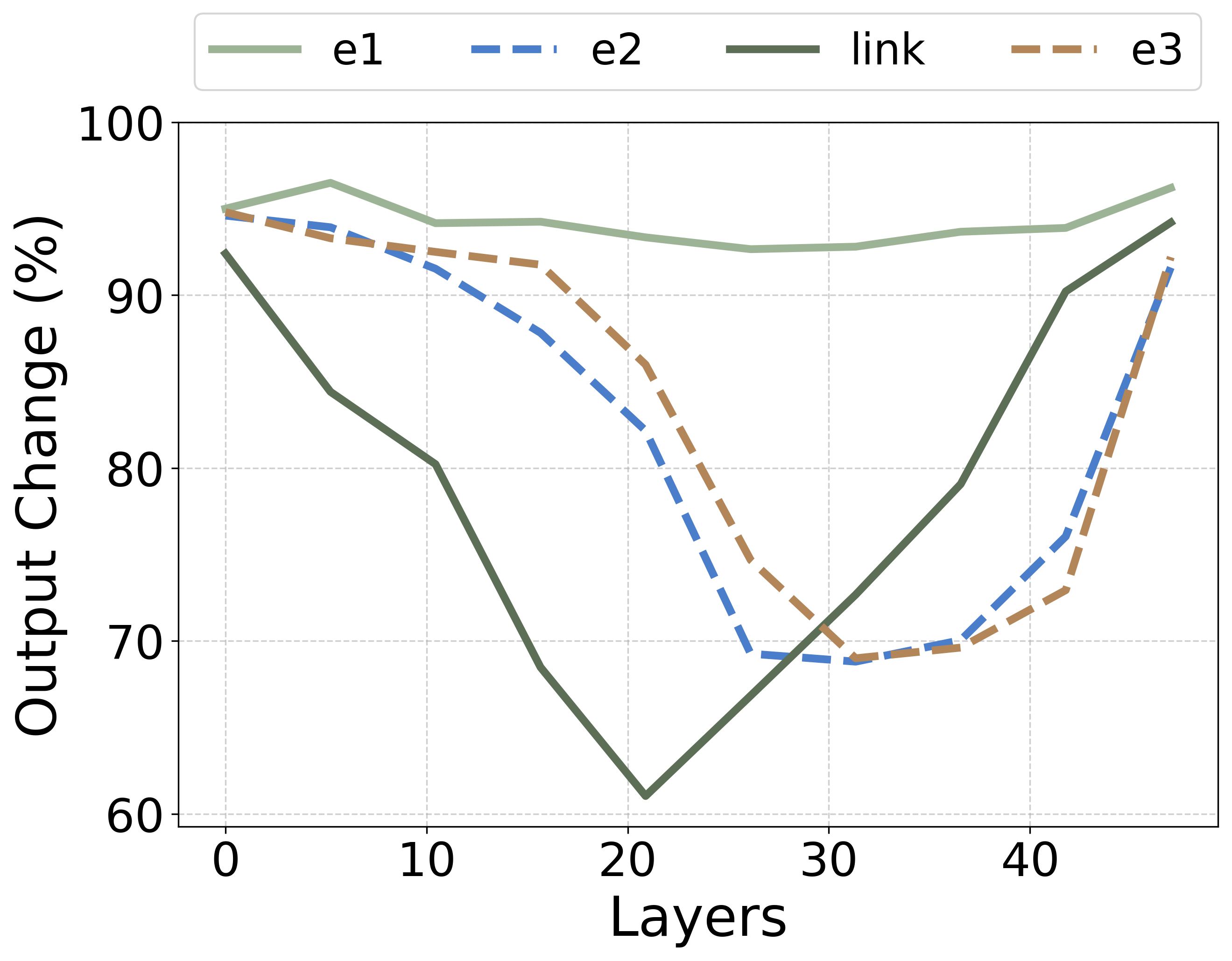}
    }
    \caption{Incorrect case} 
    \label{fig:curved_4_2_b}
\end{subfigure}

\caption{
Results of applying attention knockout to different positions on Qwen2.5-14B. Mid-upper layers of $e_2$ and $e_3$ are critical for answer resolution in both correct and incorrect cases. In incorrect cases, information from the link strongly influences model output, suggesting that the link may contribute to reasoning failures.
}
\label{fig:curved_4_2}
\end{figure*}
\subsection{Dataset Construction}
\label{subsec:dataset_construction_exp_setup}
For proportional analogies, we manually construct a test set that contains both correct and incorrect analogies for each model.
We begin by retrieving entity pairs from AnalogyKB~\cite{yuan-etal-2024-analogykb}, a million-scale analogy knowledge base that contains entity pairs of the same relation~\footnote{We use the Wikidata subset.}. 
Next, to ensure a clear distinction between correct and incorrect cases for evaluation, we manually filter out relations that can lead to multiple answers (e.g., ``\textit{interested in}") or change over time (e.g., ``\textit{head of state}"). Finally, we iteratively combine different entity pairs ($e_1$-$e_2$, $e_3$-$e_4$) that share the same relation, generating a total of 50k analogies to be used for evaluation.

In the evaluation phase, we set up a series of additional filters to confine our experiments to analogical reasoning. First, we ensure that each model is equipped with the necessary knowledge. Formally, for each ($e_i$, $e_j$) pair, we check whether models can predict $e_j$ given $e_i$ and the relation. As an illustrative example, for the analogy ``\textit{Persuasion is to Jane Austen as 1984 is to George Orwell}", we construct two queries with the relation as follows: ``\textit{The author of Persuasion is}" and ``\textit{The author of 1984 is}". If a model fails to answer both queries correctly, we exclude the analogy, as we cannot determine whether the incorrect predictions stem from incorrect analogical reasoning or from a lack of prior knowledge.
Second, we prevent models from relying on reasoning shortcuts~\cite{xu-etal-2022-model, wang-etal-2023-causal}. We define reasoning shortcuts as instances where models return the correct answer without $e_2$ or ``$e_1$ is to $e_2$". For example, we construct two queries as follows: ``\textit{Persuasion is to as 1984 is to}" and ``\textit{1984 is to}". If the model correctly predicts ``\textit{George Orwell}" in these cases, this suggests that the answer entity is strongly correlated with $e_3$, bypassing the need to perform analogical reasoning. In such cases, we exclude the analogy to ensure that models are genuinely engaging in relational reasoning rather than leveraging direct associations.
We sample 500 analogies each from the remaining collection of correct and incorrect cases for our experiments.

For story analogies, we use the StoryAnalogy~\cite{jiayang-etal-2023-storyanalogy} dataset, which contains 360 multiple-choice questions. Each question involves selecting the target story that is analogous to a given source story. The incorrect options originally consist of two randomly selected stories and one distractor story with high nounal similarity to the source story. To focus our analysis on structural alignment in the presence of surface-level distractors, we discard the random options and adopt a two-option format. To minimize positional bias, we present each question twice with reversed indices and consider a response correct only if the model selects the target story in both trials. We report detailed statistics for both datasets in the Appendix. 

\subsection{Models}
\label{subsec:models_exp_setup}

For proportional analogies, presented as a simple next-token prediction task, we investigate the following open-source models: Llama-2-13b~\cite{touvron2023llama}, Gemma-7B~\cite{team2024gemma}, and Qwen2.5-14B~\cite{yang2024qwen2}.
For story analogies, we use instruction-tuned models that demonstrate sufficient performance for analysis: Llama-2-13b-chat, Gemma-2-9B-it~\cite{gemmateam2024gemma2improvingopen}, and Qwen2.5-14B-Instruct. We mainly report results for Qwen2.5-14B models, as they exhibit representative behavior, and provide results for other models in the Appendix.


\subsection{Implementation Details}
\label{subsec:implementation_details}
For all experiments, we use two Nvidia A100 GPUs with 80GB memory. Our code is written in PyTorch (v2.3.1) and HuggingFace (v4.44.2). We report results on each model run by adopting greedy decoding to ensure reproducibility. 
\section{Information Flow in Analogical Reasoning}
\label{section:rq_1}



\begin{figure*}[t!]
\centering
\begin{subfigure}[b]{0.49\textwidth}
    \centering
    \resizebox{0.8\textwidth}{!}{ 
        \includegraphics{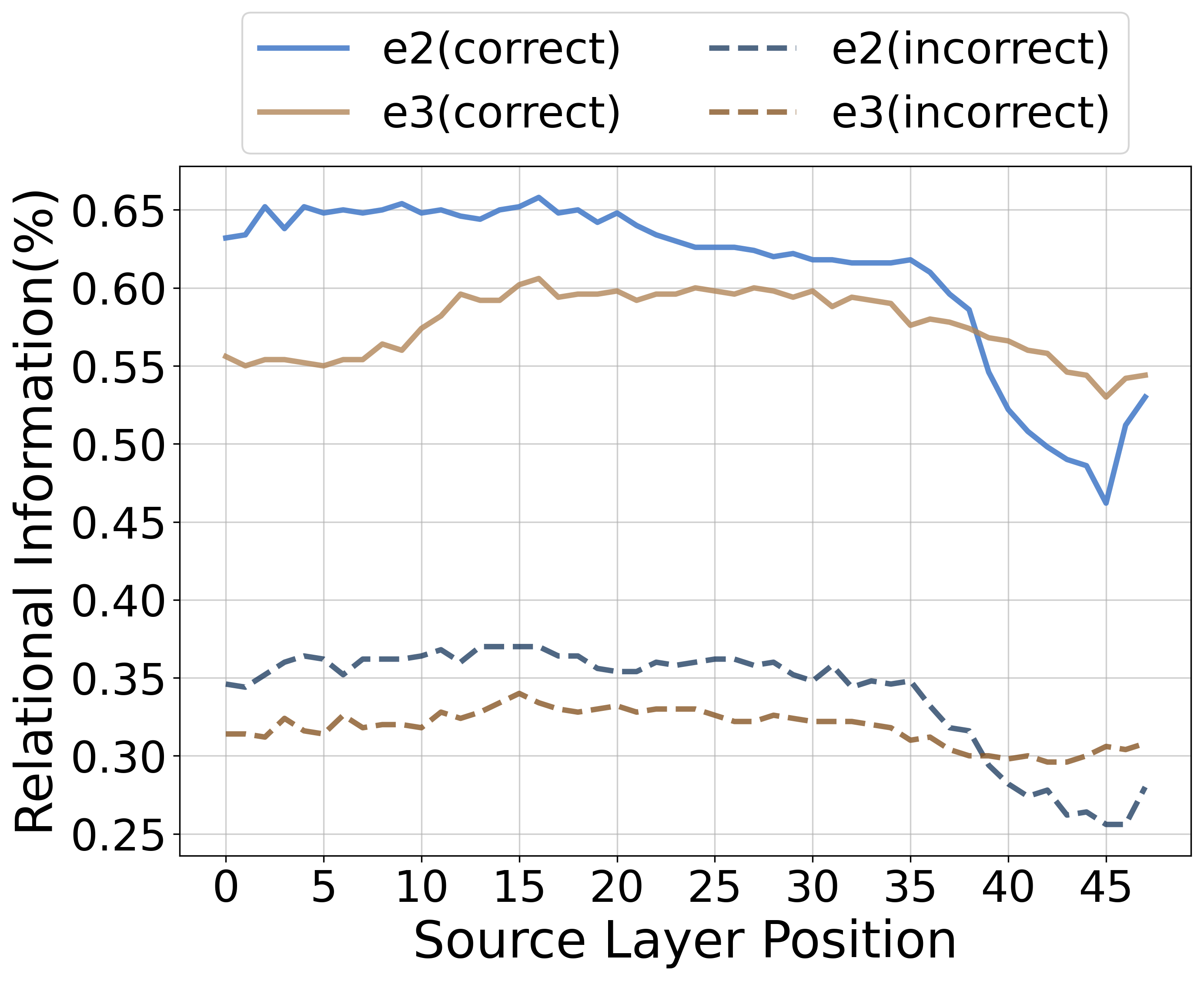}
    }
    \caption{Relational information across layers} 
    \label{fig:curved_5_2_a}
\end{subfigure}
\hfill
\begin{subfigure}[b]{0.49\textwidth}
    \centering
    \resizebox{0.8\textwidth}{!}{ 
        \includegraphics{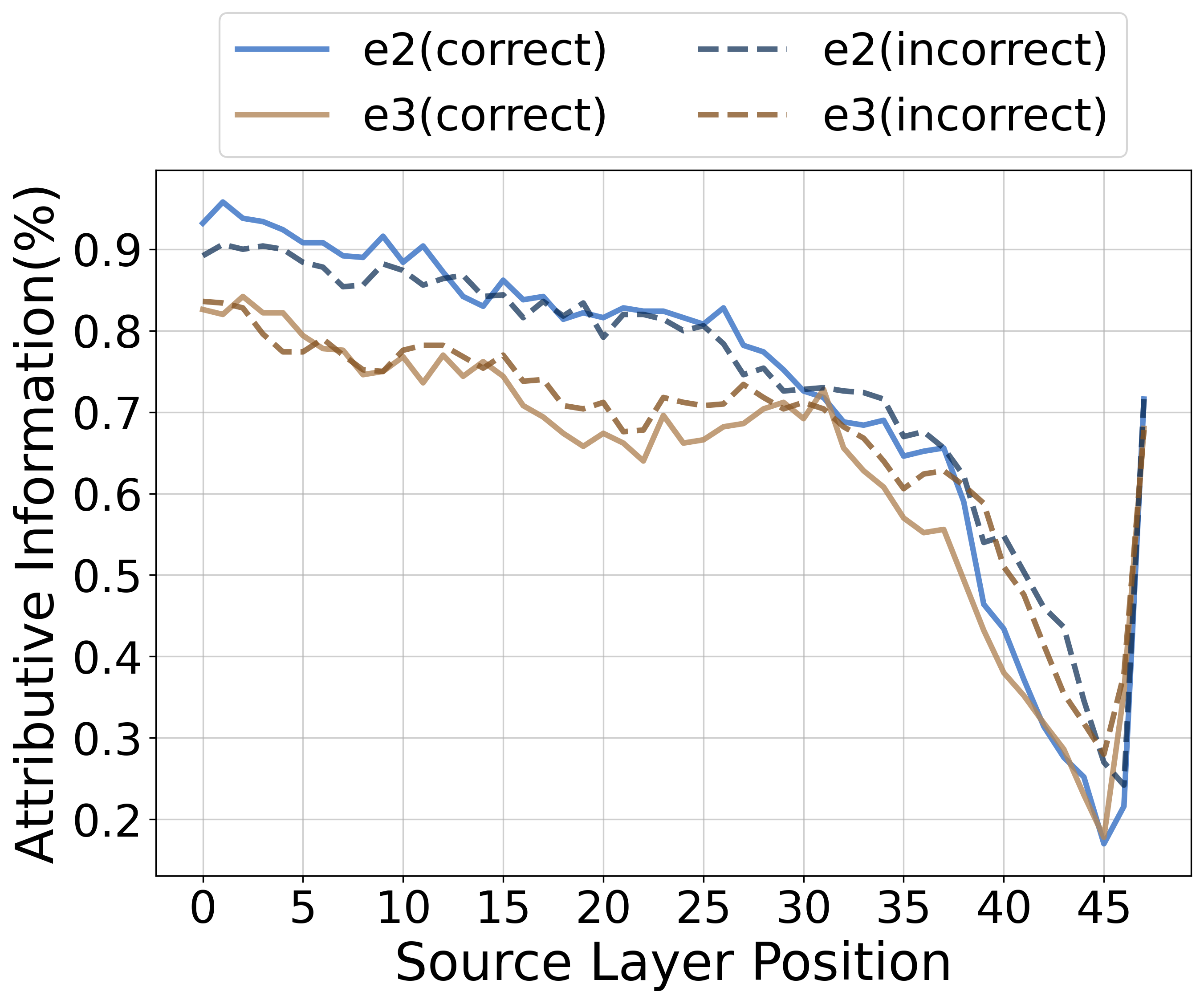}
    }
    \caption{Attributive information across layers} 
    \label{fig:curved_5_2_b}
\end{subfigure}

\caption{
Proportion of cases where relational or attributive information is successfully decoded using Patchscopes. Attributive information persists across mid-upper layers regardless of correctness, while relational information shows a sharp decline in incorrect cases. This underscores the critical role of relational information in accurate answer resolution.
}
\label{fig:curved_5_2}
\end{figure*}

The most fundamental process of analogical reasoning involves encoding the elements of an analogy, identifying the relationship between them, and applying that relationship to a target element~\cite{french2002computational, gentner2011computational}. In this section, we investigate whether this process holds true for LLMs as well using proportional analogies.

\subsection{Methods}
\label{subsec:methods_rq_1_1}
We first apply attention knockout to identify positions that are critical for resolving the answer. We focus on four positions that precede the resolution token: $e_1$, $e_2$, link, and $e_3$.
For correct cases, we report the accuracy of the generated response. For incorrect cases, we check whether the knockout results in a change in the generated text to assess the impact of blocked layers~\cite{biran-etal-2024-hopping}.
We keep a window of $k$ layers around each layer to account for information that propagates across multiple layers~\cite{geva-etal-2023-dissecting}, where $k$ is set to one-fifth of the total number of layers. 

Next, to analyze what information is encoded in the hidden representations of these positions, we first categorize information into two types: \textit{attributive} and \textit{relational} information.
Attributive information reflects how well the representation captures the inherent attributes of an entity, while relational information indicates whether the representation encodes the relation.
To analyze attributive information within each entity, we employ Patchscopes and use the same target prompt used in Section~\ref{subsec:mech_interp_preliminaries} to obtain descriptions of hidden representations in natural language. Next, to check whether each description involves the correct attributes, we take inspiration from~\citet{geva-etal-2023-dissecting} and construct a set of tokens highly related to the entity of interest. Specifically, for each entity, we retrieve 100 paragraphs from Wikipedia~\footnote{We use the dump from December 20, 2024.} using BM25~\cite{inproceedings}, and extract related entities using en\_core\_web\_trf~\cite{spacy2}. 
We consider a hidden representation to have encoded attributive information if the corresponding description contains one or more entities related to the entity of interest.

For relational information, our goal is to inspect whether a specific entity encodes the correct relation. To achieve this, we design three target prompts for each entity, encouraging models to explicitly output the encoded relation while considering their respective positions. 
For $e_2$, we use the following prompt: ``\texttt{Japan is to Tokyo: capital of, Theory of Evolution is to Charles Darwin: founder of, Peace is to olive branch: symbol of, \{\} is to x}", where curly brackets are replaced with $e_1$. 
Similarly, for $e_3$, we use the same exemplars but replace the final phrase with ``\texttt{x is to \{\}}", where curly brackets are replaced with $e_4$. 
Finally, for the resolution token (``to"), we use ``\texttt{\{\} is x}" for the final phrase. Note that we use two prompts, where curly brackets are replaced with either $e_3$ or $e_4$. 
We consider a hidden representation to have encoded relational information if the corresponding description contains the correct relation. 
We provide example descriptions generated from each custom prompt in the Appendix.

\subsection{Results}
\label{subsec:results_rq_1_1}


Figure~\ref{fig:curved_4_2} shows the results of applying attention knockout to different positions preceding the resolution token, from which we identify three notable patterns.
First, for both correct and incorrect cases, blocking attention edges from the resolution token to $e_1$ has little impact on model performance or generation. This indicates that $e_1$ plays a limited role in retaining information that is essential within the first pair.
Second, blocking attention edges to either $e_2$ or $e_3$ results in noticeable performance drops or fluctuations in generation, mainly around the mid-upper layers. This suggests that information propagating directly from $e_2$ and $e_3$ has a strong influence on model behavior, with information from $e_2$ propagating in slightly earlier layers than that from $e_3$.  
Third, information propagating from the link heavily affects model generations in incorrect cases, particularly in the early to middle layers. 
This either indicates an incorrect encoding of information passed to the link, or a failure of the link to transfer information to the target element. 
Based on this observation, we conduct further experiments to better understand incorrect cases in Section~\ref{section:rq_2}.  


Figure~\ref{fig:curved_5_2} displays the proportion of cases where relational and attributive information is successfully decoded from each source layer. We see that attributive information is consistently encoded within $e_2$ and $e_3$, persisting until the mid-upper layers before declining sharply in the upper layers. 
Given that we ensure models are equipped with the necessary knowledge (Section~\ref{subsec:dataset_construction_exp_setup}), we confirm that attributive information remains intact for $e_2$ and $e_3$ in both correct and incorrect cases. However, a significant gap is observed between these cases in terms of relational information. This suggests that relational information encoded in $e_2$ and $e_3$ serves as a key factor in answer resolution.
Moreover, while both types of information follow a similar trend for $e_2$, relational information in $e_3$ remains consistent up to the upper layers, implying its role in answer resolution at these layers.   


\section{Application as a Hurdle}
\label{section:rq_2}


For humans, the primary difficulty in solving analogies lies in extracting the underlying relation; once retrieved or cued, mapping it onto a new context is relatively straightforward~\cite{Kubricht2017}. In the previous section, we have identified two potential explanations for model failures: incorrect encoding of information passed to the link, or ineffective transfer of information through the link itself. In this section, we aim to deepen our understanding of how models fail at analogical reasoning, focusing on the observed influence of the link in incorrect generations and the pivotal role of $e_2$, which encodes both attributive and relational information. We begin by re-evaluating model performance when provided with the correct first pair. For cases where the model still fails, we then intervene by patching the representations of $e_2$ into the linking position to better facilitate the propagation of critical relational information.
\subsection{Methods}
\label{subsec:methods_rq_2}
\begin{table}[t]
\vspace{2mm}
\centering
\footnotesize
\resizebox{\columnwidth}{!}{
\begin{tabular}{lccc}
\toprule
\textbf{Model} & \textbf{Exp 1} & \textbf{Exp 2} & \textbf{Overall}\\
\midrule
Llama-2-13B & +32.3\% & +25.9\% & +49.8\%\\
\midrule
Gemma-7B & +38.4\% & +38.1\% & +61.9\%\\
\midrule
Qwen2.5-14B & +35.6\% & +30.5\% & +55.3\%\\
\bottomrule
\end{tabular}
}
\caption{
Results from error analysis experiments. ``Exp 1" indicates setting where we evaluate models using correct first pairs. ``Exp 2" indicates setting where we patch representations for the remaining incorrect cases.
}
\label{table:intervention1}
\end{table}
For the first experiment, we replace the first pairs of incorrect cases with those from correct cases.
To ensure a sufficient number of samples for replacement, we select three representative relations from our test set: ``\textit{official language of}", ``\textit{author of}", and ``\textit{composer of}". For each incorrect input analogy, we randomly choose a correct analogy from the same relation and swap their first pairs. We evaluate models using this newly constructed test set.
For the second experiment, we patch the hidden representations of each layer in $e_2$ to each layer in the link to see if models can benefit from directly injecting critical information encoded in $e_2$.
We report the performance improvement from the combination of layers that yields the highest gain.

\subsection{Results}
\label{subsec:results_rq_2}
Table~\ref{table:intervention1} shows the performance gains observed from each experiment. We find that model responses can be rectified by replacing the first pair in up to 38.4\% of incorrect cases. This indicates that a non-negligible portion of model errors stem from insufficient extraction of information within the first pair. This also highlights the importance of information encoded in $e_2$, as we have previously confirmed that the resolution token strongly attends to $e_2$ for answer resolution. 

Interestingly, for cases where replacing the first pairs did not result in correct answers, we observe that patching the representations of $e_2$ to the link leads to noticeable performance gains up to 38.1\%. This indicates that even if the model correctly extracts the necessary information from the first pair, the extent to which the link effectively conveys that information to subsequent positions can significantly impact model generation.
Moreover, we inspect the generation results across different layers for both $e_2$ and the link. For $e_2$, we find that patching representations up to the middle layers is mainly effective in rectifying model responses. 
Given that both relational and attributive information is strongly formed up to the mid-upper layers of $e_2$ in correct cases, we see that injecting information encoded from these regions into the link assists in propagating these information to subsequent positions. 
For the link, where patching is performed, applying the patched representation to the early layers proves to be effective, suggesting that the representation  need to pass through a certain number of layers to be properly contextualized with relational information.
\section{Structural Alignment in Analogies}
\label{section:rq_3}

A crucial aspect of analogical reasoning is the concept of structural alignment, i.e., the process of establishing a one-to-one correspondence between elements of two situations in a way that maximizes relational similarity~\cite{MARKMAN1993431,gentner2011computational}. This ability goes beyond recognizing lexically similar positions in context, and involves identifying parallels between seemingly unrelated, high-level concepts. In this section, we first analyze internal representations to determine whether the model distinguishes analogical context from lexically similar context. We then examine how structural alignment emerges across layers when the task is explicitly posed, and how this progression influences model behavior.

\subsection{Methods}
\label{subsec:methods_rq_3}
\begin{algorithm}[tb]
\caption{Mutual Alignment Score}
\textbf{Input}: Source token representations $S = \{s_1, \dots, s_m\}$, Candidate token representations $C = \{c_1, \dots, c_n\}$\\
\textbf{Output}: Mutual alignment score $M$

\begin{algorithmic}[1]
\STATE Normalize each vector in $S$ and $C$ to unit norm
\STATE Compute similarity matrix $M_{ij} = \cos(s_i, c_j)$ for all $i \in [1, m]$, $j \in [1, n]$
\STATE Initialize counter $\texttt{mutual\_matches} \leftarrow 0$
\FOR{$i = 1$ to $m$}
    \STATE $j^* \leftarrow \arg\max_j M_{ij}$ \hfill // Best-matching in $C$ for $s_i$
    \STATE $i^* \leftarrow \arg\max_{i'} M_{i'j^*}$ \hfill // Best-matching in $S$ for $c_{j^*}$
    \IF{$i^* = i$}
        \STATE $\texttt{mutual\_matches} \leftarrow \texttt{mutual\_matches} + 1$
    \ENDIF
\ENDFOR
\STATE $M \leftarrow \texttt{mutual\_matches} \,/\, \min(m, n)$
\STATE \textbf{return} $M$
\end{algorithmic}
\label{alg:mas}
\end{algorithm}

\noindent For the first experiment, we extract the source, target, and distractor stories from each sample in the StoryAnalogy dataset. We construct a probing dataset by pairing each source story with both the target and distractor stories. For each input pair, we extract the activation at the final token from every attention head in each layer, yielding a probing dataset $\{(x^{(h,\ell)}_i, y_i)\}_{i=1}^N$, where $x^{(h,\ell)}_i$ denotes the activation from head $h$ in layer $\ell$ for the $i$-th input pair. We train a binary linear classifier on these representations to assess whether analogical structure is linearly separable from lexical similarity in the model's internal representations. To ensure robust performance estimates and mitigate overfitting, we apply 5-fold cross-validation, reporting the average validation accuracy across folds as the final probe accuracy. 
For the second experiment, we assess whether structural alignment is reflected in the model’s internal geometry during analogical reasoning, and how it diverges between correct and incorrect cases in the presence of distractor stories. To this end, we define the \textit{Mutual Alignment Score} (MAS) as the proportion of mutual best matches between contextualized token representations from the source and candidate spans (Algorithm~\ref{alg:mas}), computed at each layer for both the target and distractor stories. A token pair $(s_i, c_j)$ forms a mutual best match if each is the other’s most similar token based on cosine similarity between their layer-specific representations. By computing MAS across layers, we trace the emergence of structural alignment and examine its relationship with successful analogical reasoning.

\subsection{Results}
\label{subsec:results_rq_3}
\begin{figure}[t!]
\centering
\includegraphics[width=\columnwidth]{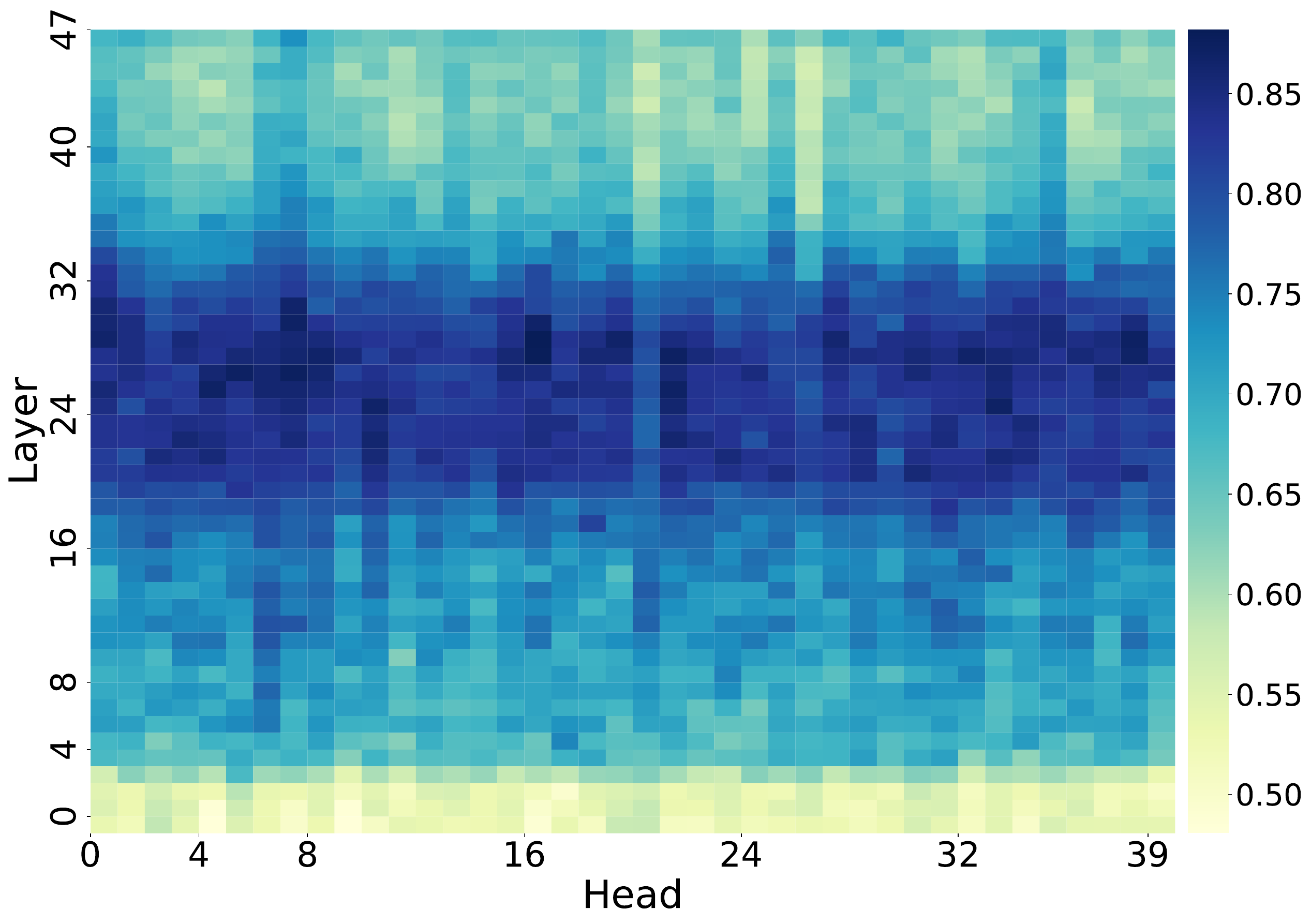}
\caption{
Linear probe accuracy across layers. High accuracy in the middle layers indicates the internal representation of analogical structure in these regions.
}
\label{fig:probing_6_1}
\end{figure}



\begin{figure}[t!]
\centering
\includegraphics[width=\columnwidth]{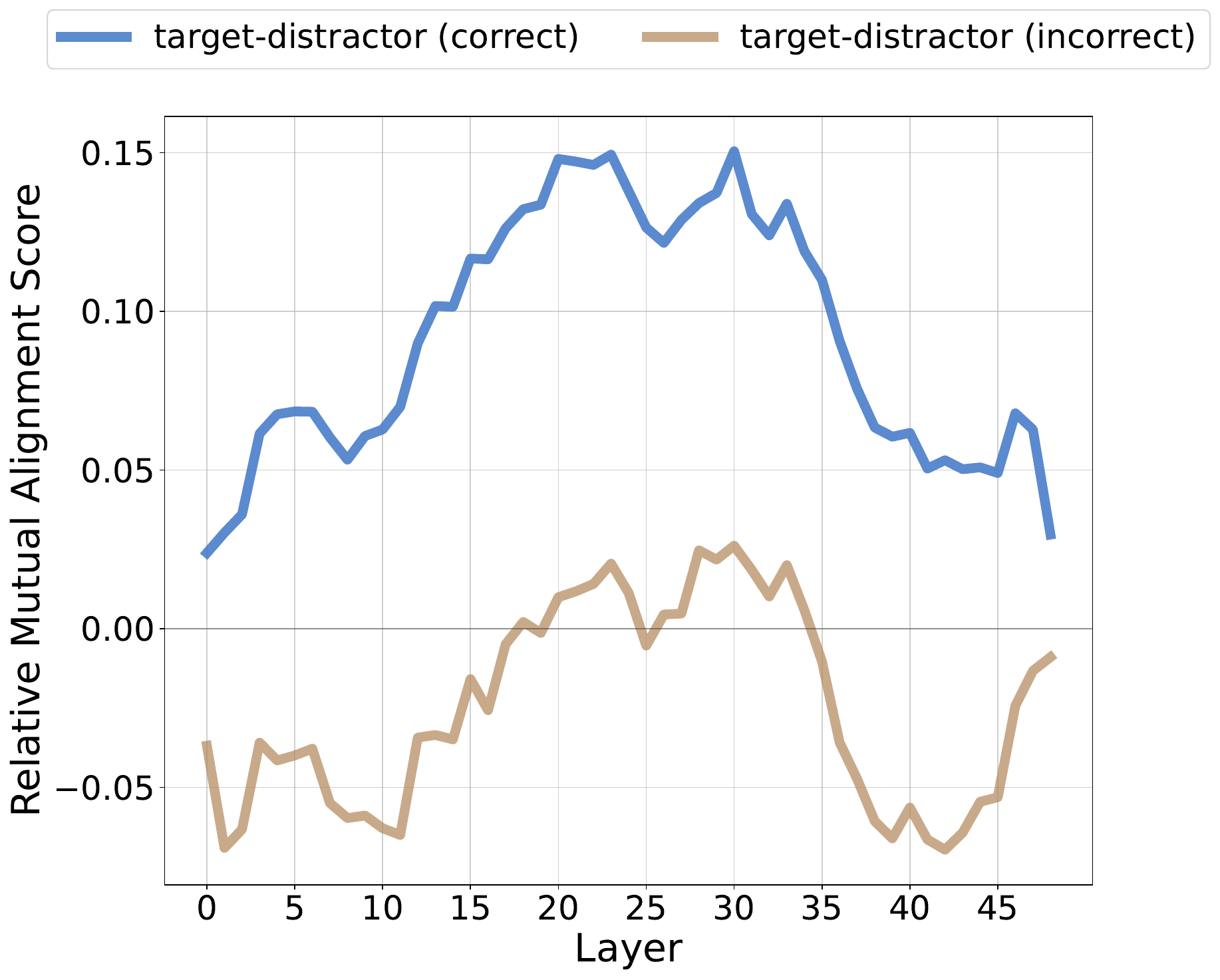}
\caption{
Relative Mutual Alignment Score (MAS) across layers, computed as the difference between the MAS of source-target pairs and source-distractor pairs.
}
\label{fig:mas_6_2}
\end{figure}
\begin{figure}[t!]
\centering
\includegraphics[width=1.0\columnwidth]{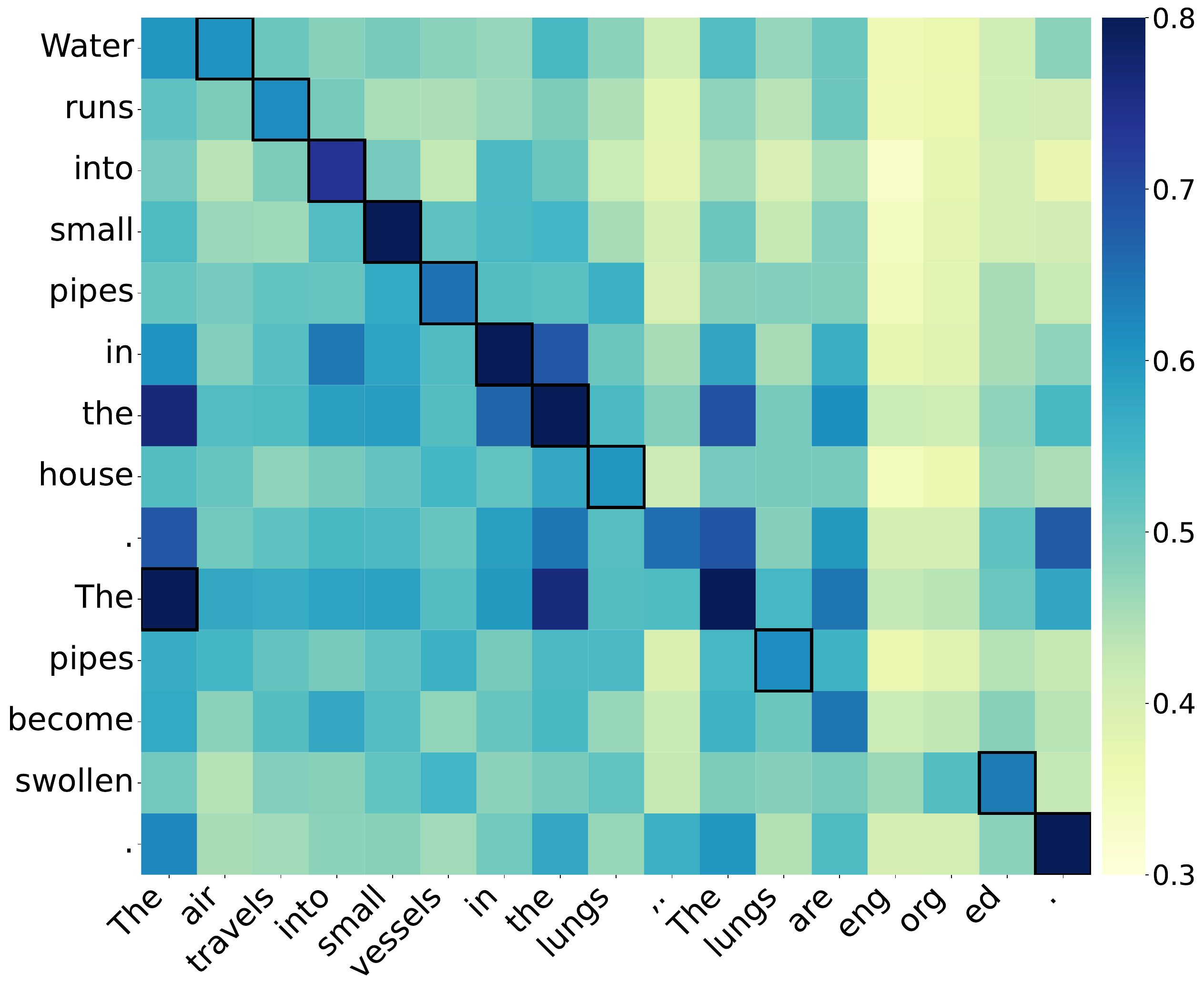}
\caption{
Sample heatmap of average similarity scores across layers between source and target stories. Black boxes indicate mutual best matches. Analogous token pairs (e.g., \textit{Water-air}, \textit{house-lungs}) form mutual best matches with high similarity scores, despite surface-level disparities.
}
\label{fig:heatmap_6_2}
\end{figure}

Figure~\ref{fig:probing_6_1} presents linear probe accuracies for distinguishing analogical from lexically similar stories across all layers of the model. The heatmap reveals a clear progression in representational quality across depth. Early to middle layers begin to show accuracies above chance, suggesting an initial emergence of analogical structure at relatively shallow depths.
A marked increase in probe accuracy extends through the middle layers, with layers 20 through 30 showing an average accuracy of 82.9\%.
This pattern indicates that analogical distinctions are not immediately encoded at the input level but instead develop gradually across layers, reaching maximal discriminability in the middle layers of the model.
These findings imply that models develop an internal representation of analogical structure that becomes linearly separable from lexical similarity as processing deepens.

Figure~\ref{fig:mas_6_2} shows the relative MAS between the source and target stories versus the source and distractor stories, measured as the difference in MAS across layers.
For correct cases, the MAS between source and target stories consistently exceeds that between source and distractor stories, suggesting that models encode deeper structural alignment beyond surface-level lexical cues. This is especially notable given that target stories are designed to have minimal entity overlap with the source, indicating that models are capturing underlying relational structure, similar to how humans seek one-to-one alignments that maximize relational similarity during analogical reasoning.

The relative gap peaks in the middle layers, suggesting that structural alignment between correct analogical pairs is strongest at these depths. This aligns with our probing results, which show that analogical distinctions become most linearly separable in these layers. In contrast, for incorrect cases, the model constructs stronger alignment between source and distractor stories across most layers, with only a slight preference for source–target alignment in the middle layers. The overall gap is much less pronounced than in correct cases, indicating that the model fails to reliably identify the intended analogical structure.


Overall, these results indicate that successful analogical reasoning in the model is strongly associated with higher token-level structural alignment between source and target stories. In contrast, incorrect cases exhibit a much smaller alignment gap, with distractors often receiving greater alignment, suggesting that the model fails to clearly differentiate the intended relational structure. This highlights structural alignment as a key internal signal for analogical success and reveals the model’s vulnerability to surface-level interference when the relational mapping is not robustly encoded.
\section{Conclusion}
\label{section:conclusion}


In this work, we study the internal mechanisms of LLMs in analogical reasoning. Using proportional analogies, we find that correct reasoning is associated with the encoding of abstract relational information in the mid-upper layers. While models are capable of abstracting these relations, we find that applying them remains a major bottleneck. By patching the representation of the second entity into the link, we uncover the link’s role in transferring relational information to downstream positions, and show that failure at this stage leads to incorrect generations. Finally, our analysis of story analogies shows that successful reasoning aligns with strong structural mapping between source and target stories, while failures often reflect weak or distractor-biased alignment. Overall, our work paves the way for future research into understanding and improving the analogical reasoning capabilities of LLMs.

\section*{Acknowledgments}
We thank Minbyul Jeong, Hyeon Hwang, and Yein Park for their invaluable feedback on this work. This research was supported by the National Research Foundation of Korea (NRF-2023R1A2C3004176), the Ministry of Health \& Welfare, Republic of Korea (HR20C002103), the Ministry of Science and ICT (MSIT) (RS-2023-00262002), the ICT Creative Consilience program through the Institute of
Information \& Communications Technology Planning \& Evaluation (IITP) grant funded by the MSIT (IITP-2025-RS-2020-II201819), and the Culture, Sports and Tourism R\&D Program through the Korea Creative Content Agency (KOCCA) grant funded by the Ministry of Culture, Sports and Tourism (MCST) in 2023 (Project Name: Development of storytelling AI technology for cultural heritage tailored to the various interests of users, Project Number: RS-2023-00220195, Contribution Rate: 100\%).

\bibliography{aaai2026}

@inproceedings{yuan-etal-2024-analogykb,
    title = "{ANALOGYKB}: Unlocking Analogical Reasoning of Language Models with A Million-scale Knowledge Base",
    author = "Yuan, Siyu  and
      Chen, Jiangjie  and
      Sun, Changzhi  and
      Liang, Jiaqing  and
      Xiao, Yanghua  and
      Yang, Deqing",
    editor = "Ku, Lun-Wei  and
      Martins, Andre  and
      Srikumar, Vivek",
    booktitle = "Proceedings of the 62nd Annual Meeting of the Association for Computational Linguistics (Volume 1: Long Papers)",
    month = aug,
    year = "2024",
    address = "Bangkok, Thailand",
    publisher = "Association for Computational Linguistics",
    url = "https://aclanthology.org/2024.acl-long.68/",
    doi = "10.18653/v1/2024.acl-long.68",
    pages = "1249--1265"
}

@inproceedings{wang-etal-2023-causal,
    title = "A Causal View of Entity Bias in (Large) Language Models",
    author = "Wang, Fei  and
      Mo, Wenjie  and
      Wang, Yiwei  and
      Zhou, Wenxuan  and
      Chen, Muhao",
    editor = "Bouamor, Houda  and
      Pino, Juan  and
      Bali, Kalika",
    booktitle = "Findings of the Association for Computational Linguistics: EMNLP 2023",
    month = dec,
    year = "2023",
    address = "Singapore",
    publisher = "Association for Computational Linguistics",
    url = "https://aclanthology.org/2023.findings-emnlp.1013/",
    doi = "10.18653/v1/2023.findings-emnlp.1013",
    pages = "15173--15184"
}

@inproceedings{xu-etal-2022-model,
    title = "Does Your Model Classify Entities Reasonably? Diagnosing and Mitigating Spurious Correlations in Entity Typing",
    author = "Xu, Nan  and
      Wang, Fei  and
      Li, Bangzheng  and
      Dong, Mingtao  and
      Chen, Muhao",
    editor = "Goldberg, Yoav  and
      Kozareva, Zornitsa  and
      Zhang, Yue",
    booktitle = "Proceedings of the 2022 Conference on Empirical Methods in Natural Language Processing",
    month = dec,
    year = "2022",
    address = "Abu Dhabi, United Arab Emirates",
    publisher = "Association for Computational Linguistics",
    url = "https://aclanthology.org/2022.emnlp-main.592/",
    doi = "10.18653/v1/2022.emnlp-main.592",
    pages = "8642--8658"
}

@article{touvron2023llama,
  title={Llama 2: Open foundation and fine-tuned chat models},
  author={Touvron, Hugo and Martin, Louis and Stone, Kevin and Albert, Peter and Almahairi, Amjad and Babaei, Yasmine and Bashlykov, Nikolay and Batra, Soumya and Bhargava, Prajjwal and Bhosale, Shruti and others},
  journal={arXiv preprint arXiv:2307.09288},
  year={2023}
}

@article{team2024gemma,
  title={Gemma: Open models based on gemini research and technology},
  author={Team, Gemma and Mesnard, Thomas and Hardin, Cassidy and Dadashi, Robert and Bhupatiraju, Surya and Pathak, Shreya and Sifre, Laurent and Rivi{\`e}re, Morgane and Kale, Mihir Sanjay and Love, Juliette and others},
  journal={arXiv preprint arXiv:2403.08295},
  year={2024}
}

@misc{gemmateam2024gemma2improvingopen,
      title={Gemma 2: Improving Open Language Models at a Practical Size}, 
      author={Gemma Team and Morgane Riviere and Shreya Pathak and Pier Giuseppe Sessa and Cassidy Hardin and Surya Bhupatiraju and Léonard Hussenot and Thomas Mesnard and Bobak Shahriari and Alexandre Ramé and Johan Ferret and Peter Liu and Pouya Tafti and Abe Friesen and Michelle Casbon and Sabela Ramos and Ravin Kumar and Charline Le Lan and Sammy Jerome and Anton Tsitsulin and Nino Vieillard and Piotr Stanczyk and Sertan Girgin and Nikola Momchev and Matt Hoffman and Shantanu Thakoor and Jean-Bastien Grill and Behnam Neyshabur and Olivier Bachem and Alanna Walton and Aliaksei Severyn and Alicia Parrish and Aliya Ahmad and Allen Hutchison and Alvin Abdagic and Amanda Carl and Amy Shen and Andy Brock and Andy Coenen and Anthony Laforge and Antonia Paterson and Ben Bastian and Bilal Piot and Bo Wu and Brandon Royal and Charlie Chen and Chintu Kumar and Chris Perry and Chris Welty and Christopher A. Choquette-Choo and Danila Sinopalnikov and David Weinberger and Dimple Vijaykumar and Dominika Rogozińska and Dustin Herbison and Elisa Bandy and Emma Wang and Eric Noland and Erica Moreira and Evan Senter and Evgenii Eltyshev and Francesco Visin and Gabriel Rasskin and Gary Wei and Glenn Cameron and Gus Martins and Hadi Hashemi and Hanna Klimczak-Plucińska and Harleen Batra and Harsh Dhand and Ivan Nardini and Jacinda Mein and Jack Zhou and James Svensson and Jeff Stanway and Jetha Chan and Jin Peng Zhou and Joana Carrasqueira and Joana Iljazi and Jocelyn Becker and Joe Fernandez and Joost van Amersfoort and Josh Gordon and Josh Lipschultz and Josh Newlan and Ju-yeong Ji and Kareem Mohamed and Kartikeya Badola and Kat Black and Katie Millican and Keelin McDonell and Kelvin Nguyen and Kiranbir Sodhia and Kish Greene and Lars Lowe Sjoesund and Lauren Usui and Laurent Sifre and Lena Heuermann and Leticia Lago and Lilly McNealus and Livio Baldini Soares and Logan Kilpatrick and Lucas Dixon and Luciano Martins and Machel Reid and Manvinder Singh and Mark Iverson and Martin Görner and Mat Velloso and Mateo Wirth and Matt Davidow and Matt Miller and Matthew Rahtz and Matthew Watson and Meg Risdal and Mehran Kazemi and Michael Moynihan and Ming Zhang and Minsuk Kahng and Minwoo Park and Mofi Rahman and Mohit Khatwani and Natalie Dao and Nenshad Bardoliwalla and Nesh Devanathan and Neta Dumai and Nilay Chauhan and Oscar Wahltinez and Pankil Botarda and Parker Barnes and Paul Barham and Paul Michel and Pengchong Jin and Petko Georgiev and Phil Culliton and Pradeep Kuppala and Ramona Comanescu and Ramona Merhej and Reena Jana and Reza Ardeshir Rokni and Rishabh Agarwal and Ryan Mullins and Samaneh Saadat and Sara Mc Carthy and Sarah Cogan and Sarah Perrin and Sébastien M. R. Arnold and Sebastian Krause and Shengyang Dai and Shruti Garg and Shruti Sheth and Sue Ronstrom and Susan Chan and Timothy Jordan and Ting Yu and Tom Eccles and Tom Hennigan and Tomas Kocisky and Tulsee Doshi and Vihan Jain and Vikas Yadav and Vilobh Meshram and Vishal Dharmadhikari and Warren Barkley and Wei Wei and Wenming Ye and Woohyun Han and Woosuk Kwon and Xiang Xu and Zhe Shen and Zhitao Gong and Zichuan Wei and Victor Cotruta and Phoebe Kirk and Anand Rao and Minh Giang and Ludovic Peran and Tris Warkentin and Eli Collins and Joelle Barral and Zoubin Ghahramani and Raia Hadsell and D. Sculley and Jeanine Banks and Anca Dragan and Slav Petrov and Oriol Vinyals and Jeff Dean and Demis Hassabis and Koray Kavukcuoglu and Clement Farabet and Elena Buchatskaya and Sebastian Borgeaud and Noah Fiedel and Armand Joulin and Kathleen Kenealy and Robert Dadashi and Alek Andreev},
      year={2024},
      eprint={2408.00118},
      archivePrefix={arXiv},
      primaryClass={cs.CL},
      url={https://arxiv.org/abs/2408.00118}, 
}

@article{yang2024qwen2,
  title={Qwen2. 5 technical report},
  author={Yang, An and Yang, Baosong and Zhang, Beichen and Hui, Binyuan and Zheng, Bo and Yu, Bowen and Li, Chengyuan and Liu, Dayiheng and Huang, Fei and Wei, Haoran and others},
  journal={arXiv preprint arXiv:2412.15115},
  year={2024}
}

@inproceedings{10.5555/3692070.3692690,
author = {Ghandeharioun, Asma and Caciularu, Avi and Pearce, Adam and Dixon, Lucas and Geva, Mor},
title = {Patchscopes: a unifying framework for inspecting hidden representations of language models},
year = {2024},
publisher = {JMLR.org},
booktitle = {Proceedings of the 41st International Conference on Machine Learning},
articleno = {620},
numpages = {25},
location = {Vienna, Austria},
series = {ICML'24}
}

@inproceedings{geva-etal-2023-dissecting,
    title = "Dissecting Recall of Factual Associations in Auto-Regressive Language Models",
    author = "Geva, Mor  and
      Bastings, Jasmijn  and
      Filippova, Katja  and
      Globerson, Amir",
    editor = "Bouamor, Houda  and
      Pino, Juan  and
      Bali, Kalika",
    booktitle = "Proceedings of the 2023 Conference on Empirical Methods in Natural Language Processing",
    month = dec,
    year = "2023",
    address = "Singapore",
    publisher = "Association for Computational Linguistics",
    url = "https://aclanthology.org/2023.emnlp-main.751/",
    doi = "10.18653/v1/2023.emnlp-main.751",
    pages = "12216--12235"
}

@inproceedings{inproceedings,
author = {Robertson, Stephen and Walker, Steve and Jones, Susan and Hancock-Beaulieu, Micheline and Gatford, Mike},
year = {1994},
month = {01},
pages = {0-},
title = {Okapi at TREC-3.}
}

@ARTICLE{spacy2,
   AUTHOR  = {Honnibal, Matthew AND Montani, Ines},
   TITLE   = {spaCy 2: Natural language understanding with Bloom embeddings, convolutional neural networks and incremental parsing},
   YEAR    = {2017},
   JOURNAL = {To appear}
}

@article{holyoak2001place,
  title={The place of analogy in cognition},
  author={Holyoak, K and Gentner, Dedre and Kokinov, B},
  journal={The analogical mind: Perspectives from cognitive science},
  volume={119},
  year={2001},
  publisher={MIT Press Cambridge, MA}
}

@article{hofstadter2001epilogue,
  title={Epilogue: Analogy as the core of cognition},
  author={Hofstadter, Douglas R},
  year={2001}
}

@book{hofstadter2013surfaces,
  title={Surfaces and essences: Analogy as the fuel and fire of thinking},
  author={Hofstadter, Douglas R and Sander, Emmanuel},
  year={2013},
  publisher={Basic books}
}

@article{jbp:/content/journals/10.1075/pc.4.2.12gen,
   author = "Gentner, Dedre and Markman, Arthur B.",
   title = "Keith J. Holyoak and Paul Thagard, Mental Leaps: Analogy in Creative Thought", 
   journal= "Pragmatics \&amp; Cognition",
   year = "1996",
   volume = "4",
   number = "2",
   pages = "407-409",
   doi = "https://doi.org/10.1075/pc.4.2.12gen",
   url = "https://www.jbe-platform.com/content/journals/10.1075/pc.4.2.12gen",
   publisher = "John Benjamins",
   issn = "0929-0907",
   type = "Journal Article",
  }

@article{doi:10.1080/713755671,
author = {Mark T. Keane},
title ={On Adaptation in Analogy: Tests of Pragmatic Importance and Adaptability in Analogical Problem Solving},
journal = {The Quarterly Journal of Experimental Psychology Section A},
volume = {49},
number = {4},
pages = {1062-1085},
year = {1996},
doi = {10.1080/713755671},
}

@article{brown1989two,
  title={Two traditions of analogy},
  author={Brown, William R},
  journal={Informal Logic},
  volume={11},
  number={3},
  year={1989}
}

@article{brown2020language,
  title={Language models are few-shot learners},
  author={Brown, Tom and Mann, Benjamin and Ryder, Nick and Subbiah, Melanie and Kaplan, Jared D and Dhariwal, Prafulla and Neelakantan, Arvind and Shyam, Pranav and Sastry, Girish and Askell, Amanda and others},
  journal={Advances in neural information processing systems},
  volume={33},
  pages={1877--1901},
  year={2020}
}

@article{logit_lens,
  title={Interpreting GPT: The Logit Lens},
  author={nostalgebraist},
  year={2020},
  journal={LessWrong},
  url={https://www.lesswrong.com/posts/AcKRB8wDpdaN6v6ru/interpreting-gpt-the-logit-lens}
}

@article{tuned_lens,
  title={Eliciting latent predictions from transformers with the tuned lens},
  author={Belrose, Nora and Furman, Zach and Smith, Logan and Halawi, Danny and Ostrovsky, Igor and McKinney, Lev and Biderman, Stella and Steinhardt, Jacob},
  journal={arXiv preprint arXiv:2303.08112},
  year={2023}
}

@inproceedings{future_lens,
    title = "Future Lens: Anticipating Subsequent Tokens from a Single Hidden State",
    author = "Pal, Koyena  and
      Sun, Jiuding  and
      Yuan, Andrew  and
      Wallace, Byron  and
      Bau, David",
    editor = "Jiang, Jing  and
      Reitter, David  and
      Deng, Shumin",
    booktitle = "Proceedings of the 27th Conference on Computational Natural Language Learning (CoNLL)",
    month = dec,
    year = "2023",
    address = "Singapore",
    publisher = "Association for Computational Linguistics",
    url = "https://aclanthology.org/2023.conll-1.37/",
    doi = "10.18653/v1/2023.conll-1.37",
}

@inproceedings{causal_mediation,
 author = {Vig, Jesse and Gehrmann, Sebastian and Belinkov, Yonatan and Qian, Sharon and Nevo, Daniel and Singer, Yaron and Shieber, Stuart},
 booktitle = {Advances in Neural Information Processing Systems},
 editor = {H. Larochelle and M. Ranzato and R. Hadsell and M.F. Balcan and H. Lin},
 pages = {12388--12401},
 publisher = {Curran Associates, Inc.},
 title = {Investigating Gender Bias in Language Models Using Causal Mediation Analysis},
 url = {https://proceedings.neurips.cc/paper_files/paper/2020/file/92650b2e92217715fe312e6fa7b90d82-Paper.pdf},
 volume = {33},
 year = {2020}
}

@inproceedings{
towards_activation_patching,
title={Towards Best Practices of Activation Patching in Language Models: Metrics and Methods},
author={Fred Zhang and Neel Nanda},
booktitle={The Twelfth International Conference on Learning Representations},
year={2024},
url={https://openreview.net/forum?id=Hf17y6u9BC}
}

@inproceedings{
paragraph_activation,
title={Extracting Paragraphs from {LLM} Token Activations},
author={Nicky Pochinkov and Angelo Benoit and Lovkush Agarwal and Zainab Ali Majid and Lucile Ter-Minassian},
booktitle={MINT: Foundation Model Interventions},
year={2024},
url={https://openreview.net/forum?id=4b675AHcqq}
}

@article{kojima2022large,
  title={Large language models are zero-shot reasoners},
  author={Kojima, Takeshi and Gu, Shixiang Shane and Reid, Machel and Matsuo, Yutaka and Iwasawa, Yusuke},
  journal={Advances in neural information processing systems},
  volume={35},
  pages={22199--22213},
  year={2022}
}

@article{imani2023mathprompter,
  title={Mathprompter: Mathematical reasoning using large language models},
  author={Imani, Shima and Du, Liang and Shrivastava, Harsh},
  journal={arXiv preprint arXiv:2303.05398},
  year={2023}
}

@inproceedings{yao2023react,
  title = {{ReAct}: Synergizing Reasoning and Acting in Language Models},
  author = {Yao, Shunyu and Zhao, Jeffrey and Yu, Dian and Du, Nan and Shafran, Izhak and Narasimhan, Karthik and Cao, Yuan},
  booktitle = {International Conference on Learning Representations (ICLR) },
  year = {2023},
  html = {https://arxiv.org/abs/2210.03629},
}

@inproceedings{NEURIPS2023_631bb943,
 author = {Jin, Zhijing and Chen, Yuen and Leeb, Felix and Gresele, Luigi and Kamal, Ojasv and LYU, Zhiheng and Blin, Kevin and Gonzalez Adauto, Fernando and Kleiman-Weiner, Max and Sachan, Mrinmaya and Sch\"{o}lkopf, Bernhard},
 booktitle = {Advances in Neural Information Processing Systems},
 editor = {A. Oh and T. Naumann and A. Globerson and K. Saenko and M. Hardt and S. Levine},
 pages = {31038--31065},
 publisher = {Curran Associates, Inc.},
 title = {CLadder: Assessing Causal Reasoning in Language Models},
 url = {https://proceedings.neurips.cc/paper_files/paper/2023/file/631bb9434d718ea309af82566347d607-Paper-Conference.pdf},
 volume = {36},
 year = {2023}
}

@article{bereska2024mechanistic,
  title={Mechanistic Interpretability for AI Safety--A Review},
  author={Bereska, Leonard and Gavves, Efstratios},
  journal={arXiv preprint arXiv:2404.14082},
  year={2024}
}

@inproceedings{
wang2023interpretability,
title={Interpretability in the Wild: a Circuit for Indirect Object Identification in {GPT}-2 Small},
author={Kevin Ro Wang and Alexandre Variengien and Arthur Conmy and Buck Shlegeris and Jacob Steinhardt},
booktitle={The Eleventh International Conference on Learning Representations },
year={2023},
url={https://openreview.net/forum?id=NpsVSN6o4ul}
}

@inproceedings{boteanu2015solving,
  title={Solving and explaining analogy questions using semantic networks},
  author={Boteanu, Adrian and Chernova, Sonia},
  booktitle={Proceedings of the AAAI Conference on Artificial Intelligence},
  volume={29},
  number={1},
  year={2015}
}

@inproceedings{sultan-shahaf-2022-life,
    title = "Life is a Circus and We are the Clowns: Automatically Finding Analogies between Situations and Processes",
    author = "Sultan, Oren  and
      Shahaf, Dafna",
    editor = "Goldberg, Yoav  and
      Kozareva, Zornitsa  and
      Zhang, Yue",
    booktitle = "Proceedings of the 2022 Conference on Empirical Methods in Natural Language Processing",
    month = dec,
    year = "2022",
    address = "Abu Dhabi, United Arab Emirates",
    publisher = "Association for Computational Linguistics",
    url = "https://aclanthology.org/2022.emnlp-main.232/",
    doi = "10.18653/v1/2022.emnlp-main.232",
    pages = "3547--3562"
}

@inproceedings{jiayang-etal-2023-storyanalogy,
    title = "{S}tory{A}nalogy: Deriving Story-level Analogies from Large Language Models to Unlock Analogical Understanding",
    author = "Jiayang, Cheng  and
      Qiu, Lin  and
      Chan, Tsz  and
      Fang, Tianqing  and
      Wang, Weiqi  and
      Chan, Chunkit  and
      Ru, Dongyu  and
      Guo, Qipeng  and
      Zhang, Hongming  and
      Song, Yangqiu  and
      Zhang, Yue  and
      Zhang, Zheng",
    editor = "Bouamor, Houda  and
      Pino, Juan  and
      Bali, Kalika",
    booktitle = "Proceedings of the 2023 Conference on Empirical Methods in Natural Language Processing",
    month = dec,
    year = "2023",
    address = "Singapore",
    publisher = "Association for Computational Linguistics",
    url = "https://aclanthology.org/2023.emnlp-main.706/",
    doi = "10.18653/v1/2023.emnlp-main.706",
    pages = "11518--11537"
}

@inproceedings{mikolov-etal-2013-linguistic,
    title = "Linguistic Regularities in Continuous Space Word Representations",
    author = "Mikolov, Tomas  and
      Yih, Wen-tau  and
      Zweig, Geoffrey",
    editor = "Vanderwende, Lucy  and
      Daum{\'e} III, Hal  and
      Kirchhoff, Katrin",
    booktitle = "Proceedings of the 2013 Conference of the North {A}merican Chapter of the Association for Computational Linguistics: Human Language Technologies",
    month = jun,
    year = "2013",
    address = "Atlanta, Georgia",
    publisher = "Association for Computational Linguistics",
    url = "https://aclanthology.org/N13-1090/",
    pages = "746--751"
}

@article{webb2023emergent,
  title={Emergent analogical reasoning in large language models},
  author={Webb, Taylor and Holyoak, Keith J and Lu, Hongjing},
  journal={Nature Human Behaviour},
  volume={7},
  number={9},
  pages={1526--1541},
  year={2023},
  publisher={Nature Publishing Group UK London}
}

@inproceedings{gladkova-etal-2016-analogy,
    title = "Analogy-based detection of morphological and semantic relations with word embeddings: what works and what doesn`t.",
    author = "Gladkova, Anna  and
      Drozd, Aleksandr  and
      Matsuoka, Satoshi",
    editor = "Andreas, Jacob  and
      Choi, Eunsol  and
      Lazaridou, Angeliki",
    booktitle = "Proceedings of the {NAACL} Student Research Workshop",
    month = jun,
    year = "2016",
    address = "San Diego, California",
    publisher = "Association for Computational Linguistics",
    url = "https://aclanthology.org/N16-2002/",
    doi = "10.18653/v1/N16-2002",
    pages = "8--15"
}

@article{wijesiriwardene2023analogical,
  title={ANALOGICAL--A Novel Benchmark for Long Text Analogy Evaluation in Large Language Models},
  author={Wijesiriwardene, Thilini and Wickramarachchi, Ruwan and Gajera, Bimal G and Gowaikar, Shreeyash Mukul and Gupta, Chandan and Chadha, Aman and Reganti, Aishwarya Naresh and Sheth, Amit and Das, Amitava},
  journal={arXiv preprint arXiv:2305.05050},
  year={2023}
}

@inproceedings{
yasunaga2024large,
title={Large Language Models as Analogical Reasoners},
author={Michihiro Yasunaga and Xinyun Chen and Yujia Li and Panupong Pasupat and Jure Leskovec and Percy Liang and Ed H. Chi and Denny Zhou},
booktitle={The Twelfth International Conference on Learning Representations},
year={2024},
url={https://openreview.net/forum?id=AgDICX1h50}
}

@inproceedings{ye-etal-2024-analobench,
    title = "{A}nalo{B}ench: Benchmarking the Identification of Abstract and Long-context Analogies",
    author = "Ye, Xiao  and
      Wang, Andrew  and
      Choi, Jacob  and
      Lu, Yining  and
      Sharma, Shreya  and
      Shen, Lingfeng  and
      Tiyyala, Vijay Murari  and
      Andrews, Nicholas  and
      Khashabi, Daniel",
    editor = "Al-Onaizan, Yaser  and
      Bansal, Mohit  and
      Chen, Yun-Nung",
    booktitle = "Proceedings of the 2024 Conference on Empirical Methods in Natural Language Processing",
    month = nov,
    year = "2024",
    address = "Miami, Florida, USA",
    publisher = "Association for Computational Linguistics",
    url = "https://aclanthology.org/2024.emnlp-main.725/",
    doi = "10.18653/v1/2024.emnlp-main.725",
    pages = "13060--13082"
}

@article{wijesiriwardene2024exploring,
  title={Exploring the Abilities of Large Language Models to Solve Proportional Analogies via Knowledge-Enhanced Prompting},
  author={Wijesiriwardene, Thilini and Wickramarachchi, Ruwan and Vennam, Sreeram and Jain, Vinija and Chadha, Aman and Das, Amitava and Kumaraguru, Ponnurangam and Sheth, Amit},
  journal={arXiv preprint arXiv:2412.00869},
  year={2024}
}

@article{french2002computational,
  title={The computational modeling of analogy-making},
  author={French, Robert M},
  journal={Trends in cognitive Sciences},
  volume={6},
  number={5},
  pages={200--205},
  year={2002},
  publisher={Elsevier}
}

@article{gentner2011computational,
  title={Computational models of analogy},
  author={Gentner, Dedre and Forbus, Kenneth D},
  journal={Wiley interdisciplinary reviews: cognitive science},
  volume={2},
  number={3},
  pages={266--276},
  year={2011},
  publisher={Wiley Online Library}
}

@inproceedings{biran-etal-2024-hopping,
    title = "Hopping Too Late: Exploring the Limitations of Large Language Models on Multi-Hop Queries",
    author = "Biran, Eden  and
      Gottesman, Daniela  and
      Yang, Sohee  and
      Geva, Mor  and
      Globerson, Amir",
    editor = "Al-Onaizan, Yaser  and
      Bansal, Mohit  and
      Chen, Yun-Nung",
    booktitle = "Proceedings of the 2024 Conference on Empirical Methods in Natural Language Processing",
    month = nov,
    year = "2024",
    address = "Miami, Florida, USA",
    publisher = "Association for Computational Linguistics",
    url = "https://aclanthology.org/2024.emnlp-main.781/",
    doi = "10.18653/v1/2024.emnlp-main.781",
    pages = "14113--14130"
}

@article{MARKMAN1993431,
title = {Structural Alignment during Similarity Comparisons},
journal = {Cognitive Psychology},
volume = {25},
number = {4},
pages = {431-467},
year = {1993},
issn = {0010-0285},
doi = {https://doi.org/10.1006/cogp.1993.1011},
url = {https://www.sciencedirect.com/science/article/pii/S001002858371011X},
author = {A.B. Markman and D. Gentner}
}

@inproceedings{hendel-etal-2023-context,
    title = "In-Context Learning Creates Task Vectors",
    author = "Hendel, Roee  and
      Geva, Mor  and
      Globerson, Amir",
    editor = "Bouamor, Houda  and
      Pino, Juan  and
      Bali, Kalika",
    booktitle = "Findings of the Association for Computational Linguistics: EMNLP 2023",
    month = dec,
    year = "2023",
    address = "Singapore",
    publisher = "Association for Computational Linguistics",
    url = "https://aclanthology.org/2023.findings-emnlp.624/",
    doi = "10.18653/v1/2023.findings-emnlp.624",
    pages = "9318--9333"
}

@inproceedings{
todd2024function,
title={Function Vectors in Large Language Models},
author={Eric Todd and Millicent Li and Arnab Sen Sharma and Aaron Mueller and Byron C Wallace and David Bau},
booktitle={The Twelfth International Conference on Learning Representations},
year={2024},
url={https://openreview.net/forum?id=AwyxtyMwaG}
}

@article{opielka2025analogical,
  title={Analogical reasoning inside large language models: Concept vectors and the limits of abstraction},
  author={Opie{\l}ka, Gustaw and Rosenbusch, Hannes and Stevenson, Claire E},
  journal={arXiv preprint arXiv:2503.03666},
  year={2025}
}

@misc{alain2018understandingintermediatelayersusing,
      title={Understanding intermediate layers using linear classifier probes}, 
      author={Guillaume Alain and Yoshua Bengio},
      year={2018},
      eprint={1610.01644},
      archivePrefix={arXiv},
      primaryClass={stat.ML},
      url={https://arxiv.org/abs/1610.01644}, 
}

@article{belinkov-2022-probing,
    title = "Probing Classifiers: Promises, Shortcomings, and Advances",
    author = "Belinkov, Yonatan",
    journal = "Computational Linguistics",
    volume = "48",
    number = "1",
    month = mar,
    year = "2022",
    address = "Cambridge, MA",
    publisher = "MIT Press",
    url = "https://aclanthology.org/2022.cl-1.7/",
    doi = "10.1162/coli_a_00422",
    pages = "207--219"
}

@Article{Kubricht2017,
author={Kubricht, James R.
and Lu, Hongjing
and Holyoak, Keith J.},
title={Individual differences in spontaneous analogical transfer},
journal={Memory {\&} Cognition},
year={2017},
month={May},
day={01},
volume={45},
number={4},
pages={576-588},
issn={1532-5946},
doi={10.3758/s13421-016-0687-7},
url={https://doi.org/10.3758/s13421-016-0687-7}
}

\appendix
\clearpage
\renewcommand{\thetable}{\Alph{table}}
\setcounter{table}{0}
\renewcommand{\thefigure}{\Alph{figure}}
\setcounter{figure}{0}

\section{Dataset Statistics}
In Table~\ref{table:dataset_proportional} and~\ref{table:dataset_story}, we report detailed statistics of data samples used in proportional and story analogies per model, respectively. For proportional analogies, ``Total" refers to the total number of samples after the knowledge filtering process. We then sample 500 instances from both correct and incorrect cases for analysis.
\label{appendix:dataset_statistics}

\section{Example Descriptions of Target Prompts}
In Table~\ref{table:author_of_prompts},~\ref{table:director_of_prompts}, and~\ref{table:composer_of_prompts}, we provide example descriptions generated using Patchscopes across different token positions and relations. Specifically, we apply different versions of custom prompts to encourage models to output relational information encoded in each entity position. For attributive information, we apply a fixed default prompt on each entity position, as represented in \texttt{resolution (default)}. 
\label{appendix:target_prompt_descriptions}

\section{Heatmap of Answer Resolution in Resolution Token}
\begin{figure*}[t!]
\centering
\begin{subfigure}[b]{0.3\textwidth}
    \centering
    \includegraphics[width=\textwidth]{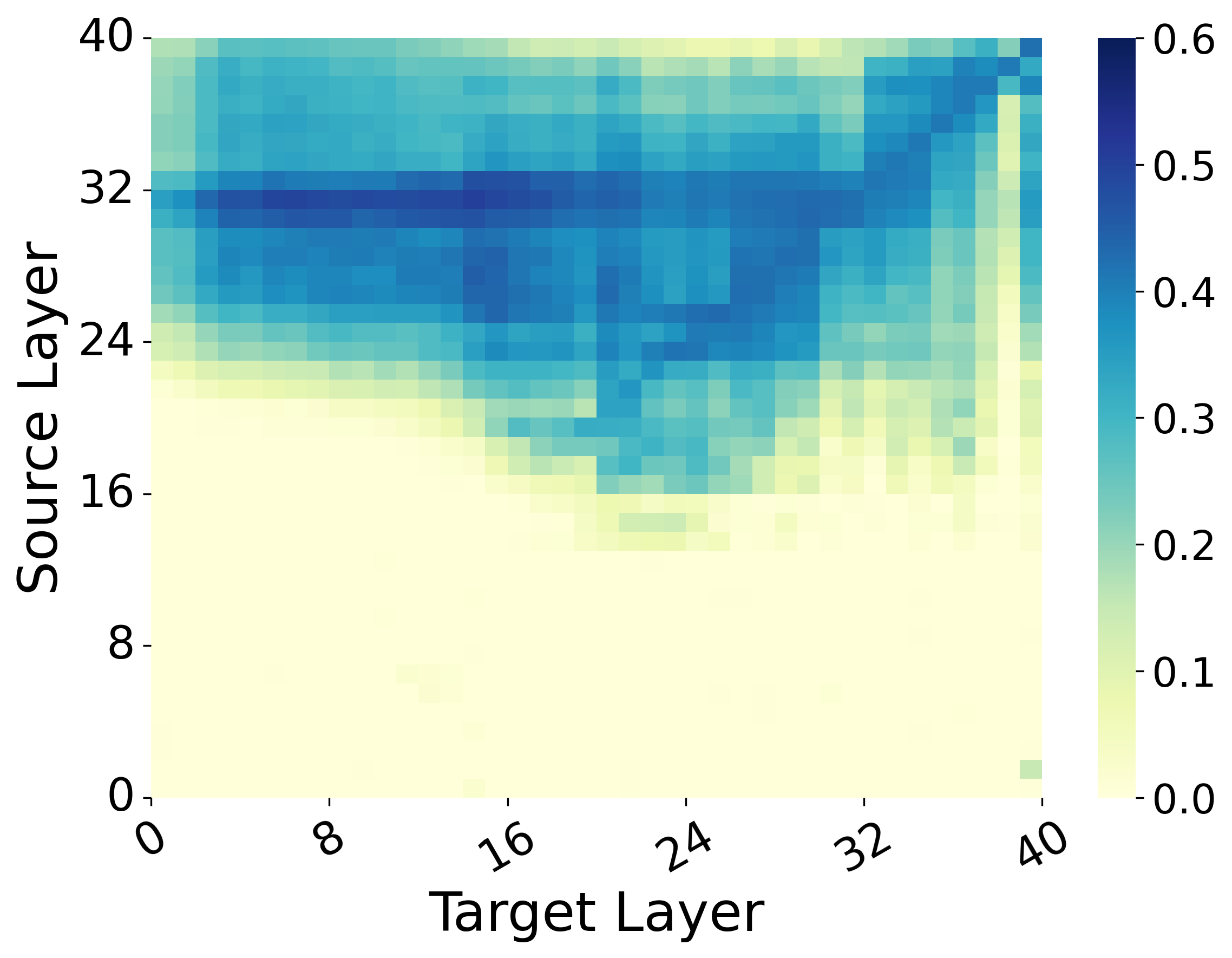}
    \caption{Llama-2-13B}
\end{subfigure}
\hfill
\begin{subfigure}[b]{0.3\textwidth}
    \centering
    \includegraphics[width=\textwidth]{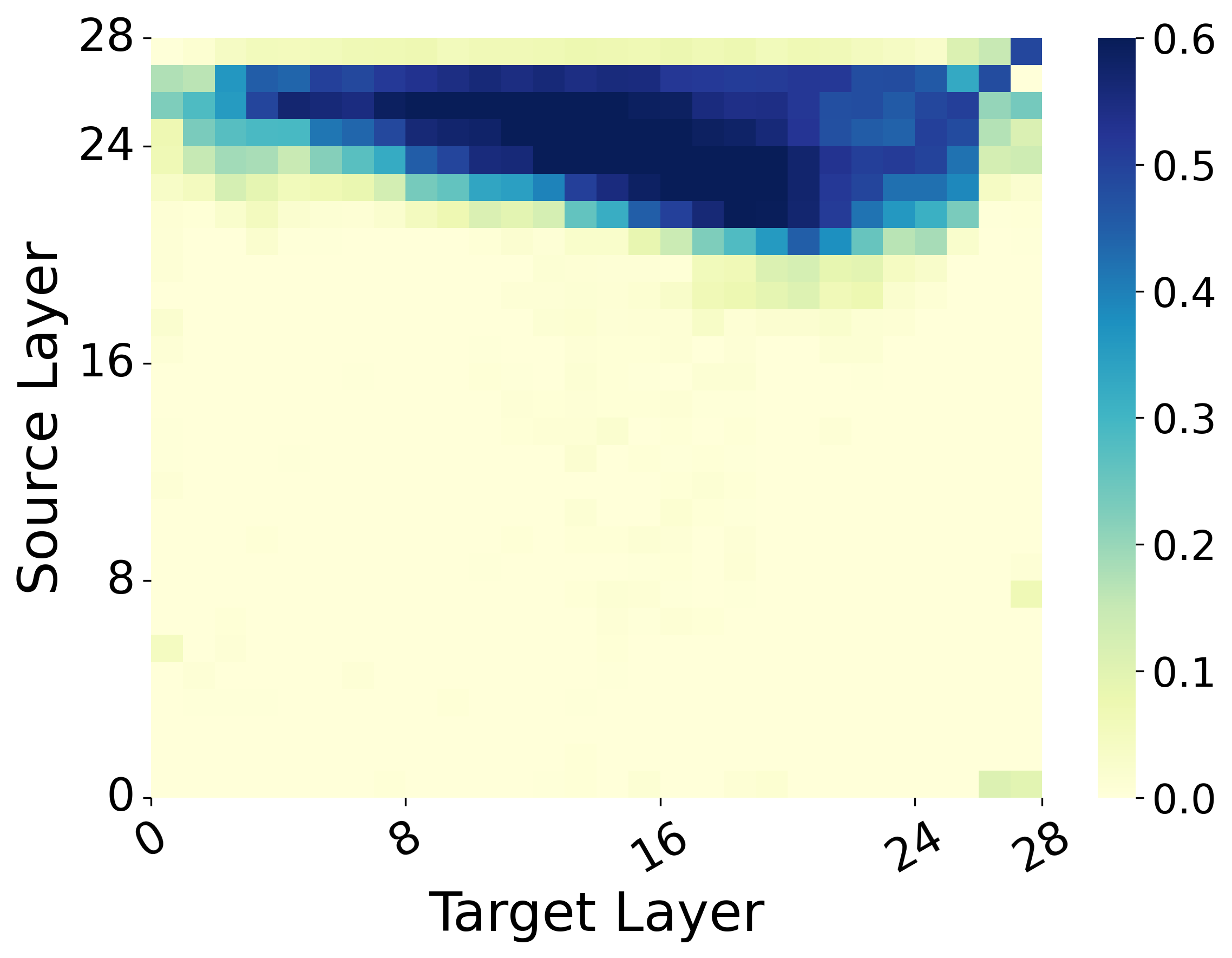}
    \caption{Gemma-7B}
\end{subfigure}
\hfill
\begin{subfigure}[b]{0.3\textwidth}
    \centering
    \includegraphics[width=\textwidth]{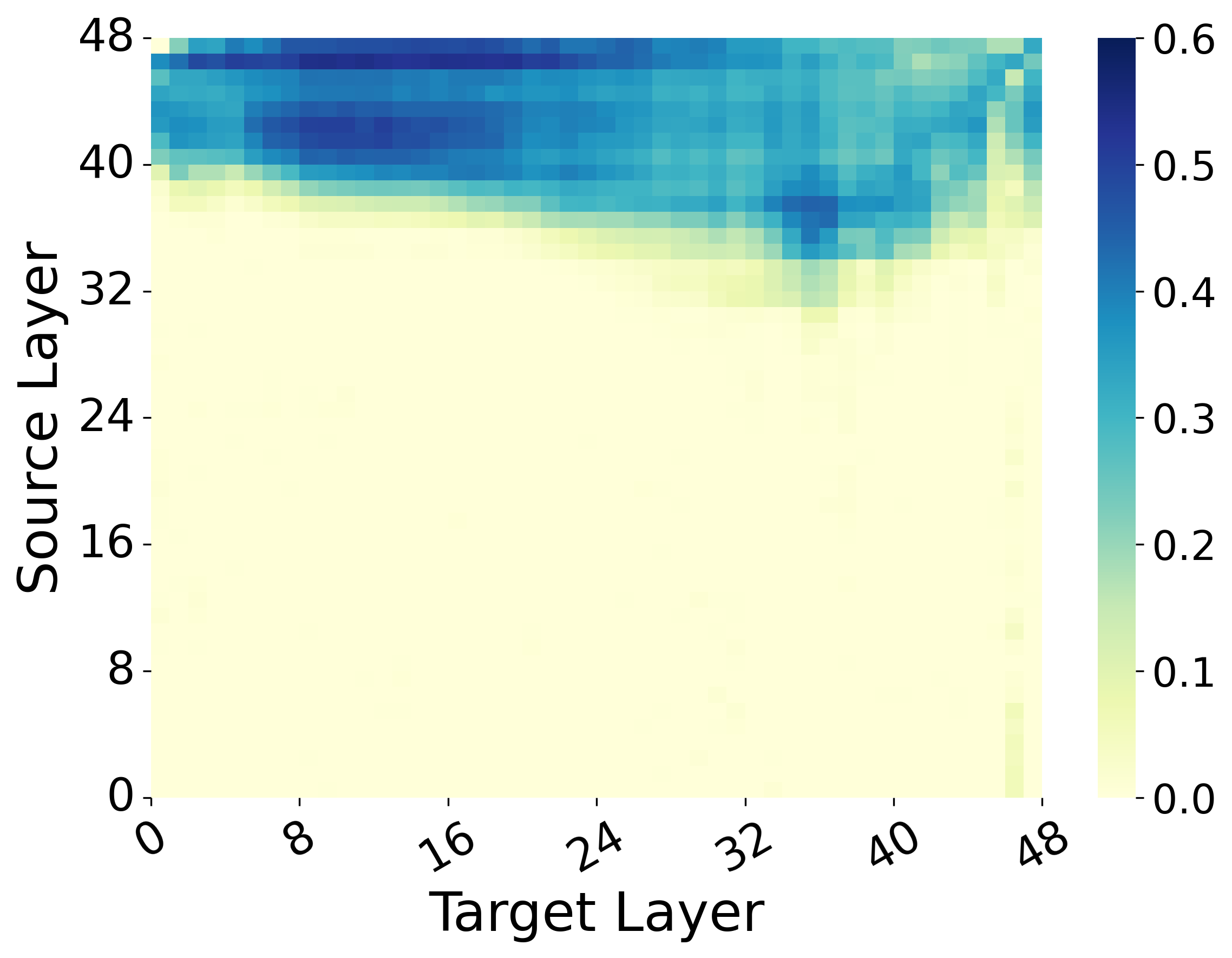}
    \caption{Qwen2.5-14B}
\end{subfigure}
\caption{Heatmap of layers in the resolution token where $e_4$ is successfully decoded using Patchscopes.}

\label{fig:heatmap_resolution_token}
\end{figure*}

Figure~\ref{fig:heatmap_resolution_token} depicts the heatmap of layers in the resolution token where $e_4$ is successfully decoded using Patchscopes for all models. We confirm that the answer is mostly resolved in the upper layers of the resolution token, and look for positions that propagate information critical for this process.
\label{appendix:heatmap_resolution_token}

\section{Attention Knockout on Different Positions}
\begin{figure*}[t!]
\centering
\begin{subfigure}[b]{0.49\textwidth}
    \centering
    \includegraphics[width=\textwidth]{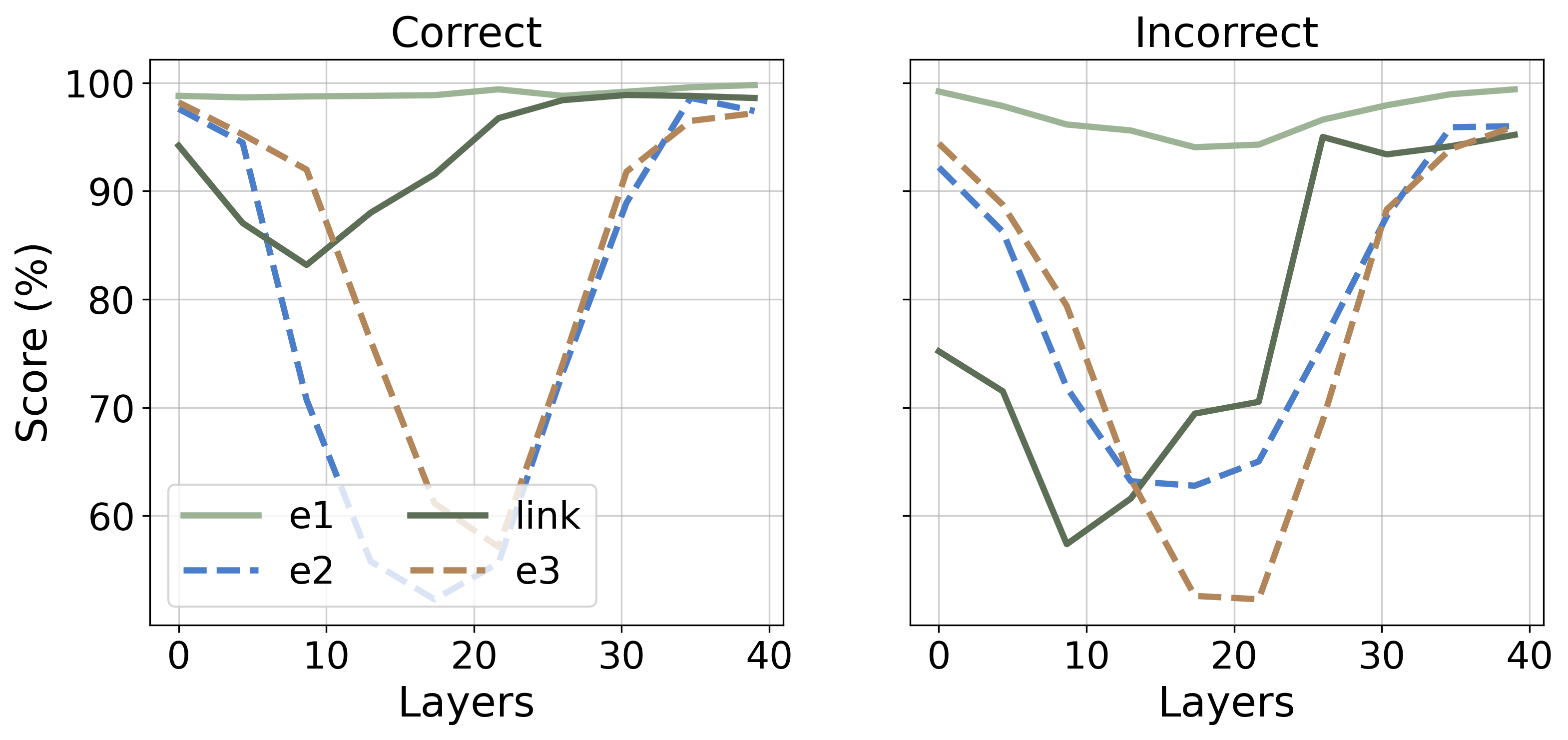}
    \caption{Llama-2-13B}
\end{subfigure}
\hfill
\begin{subfigure}[b]{0.49\textwidth}
    \centering
    \includegraphics[width=\textwidth]{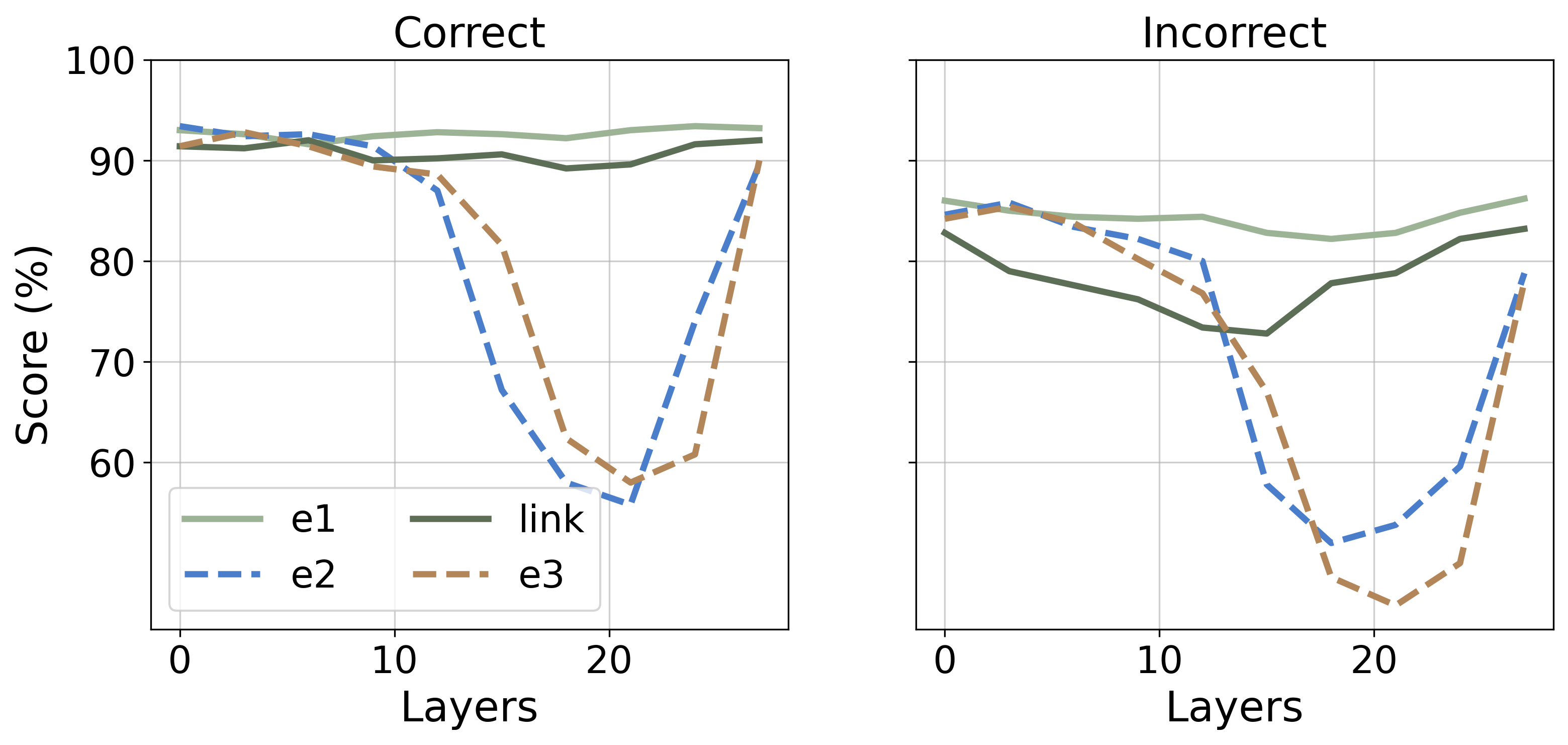}
    \caption{Gemma-7B}
\end{subfigure}
\hfill
\begin{subfigure}[b]{0.49\textwidth}
    \centering
    \includegraphics[width=\textwidth]{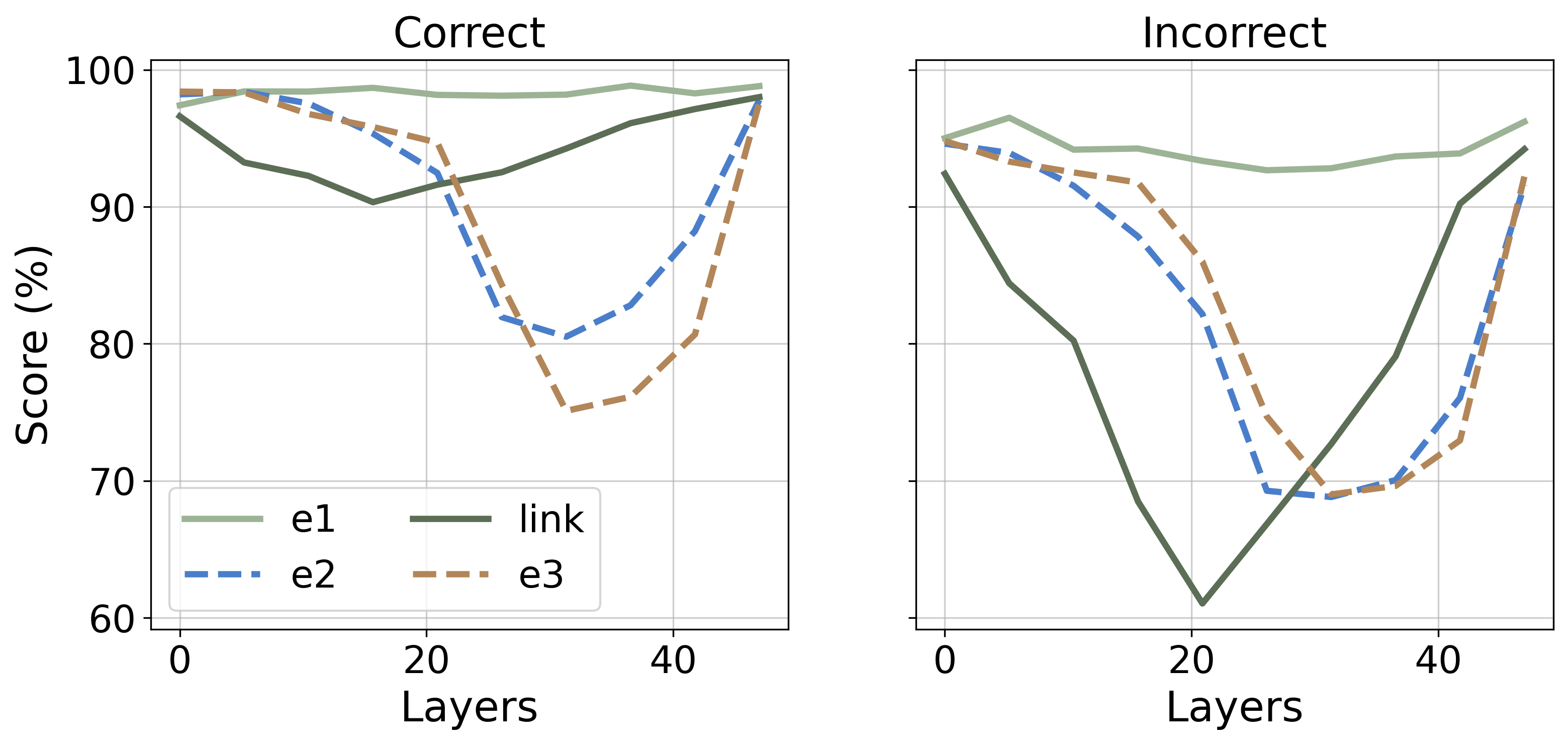}
    \caption{Qwen2.5-14B}
\end{subfigure}
\caption{Attention knockout results for all models.}
\label{fig:curved_attention_knockout}
\end{figure*}

Figure~\ref{fig:curved_attention_knockout} shows the results of applying attention knockout to different positions for all models.
\label{appendix:curved_attention_knockout}

\section{Decoding Information on Different Positions}
\begin{figure*}[t!]
\centering
\begin{subfigure}[b]{0.3\textwidth}
    \centering
    \includegraphics[width=\textwidth]{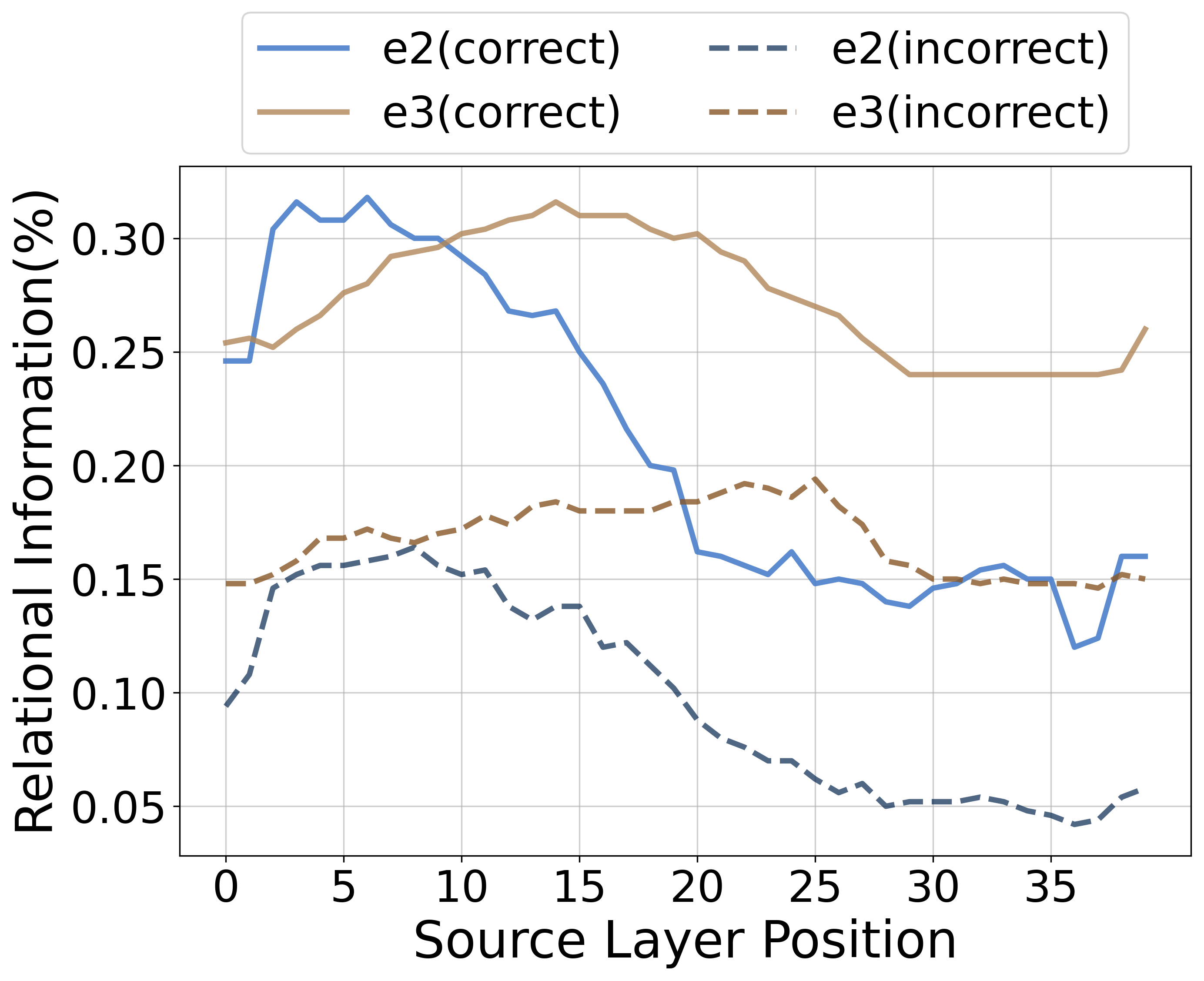}
    \caption{Llama-2-13B}
\end{subfigure}
\hfill
\begin{subfigure}[b]{0.3\textwidth}
    \centering
    \includegraphics[width=\textwidth]{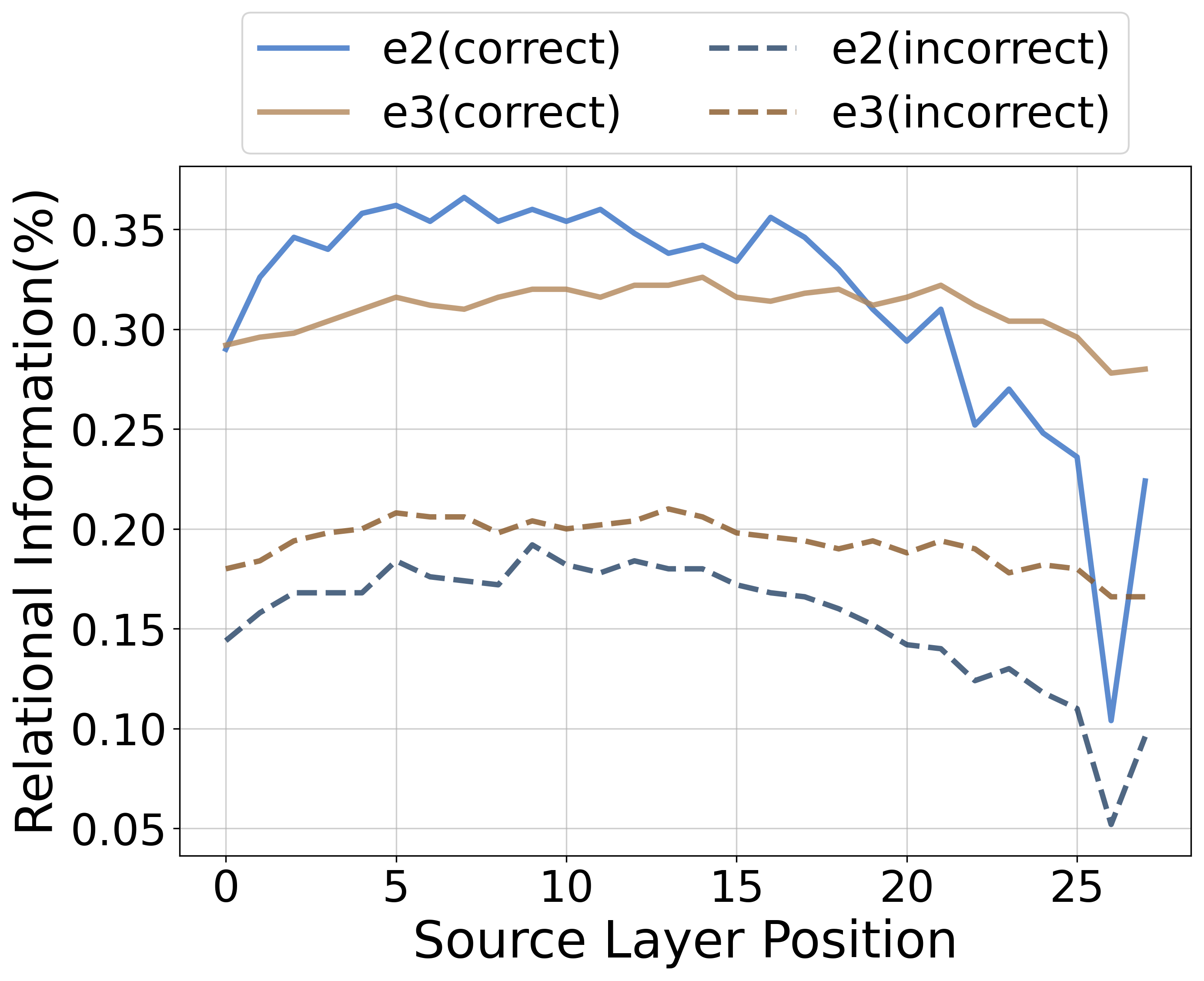}
    \caption{Gemma-7B}
\end{subfigure}
\hfill
\begin{subfigure}[b]{0.3\textwidth}
    \centering
    \includegraphics[width=\textwidth]{figures/appendix/relational_Qwen2.5-14B.png}
    \caption{Qwen2.5-14B}
\end{subfigure}
\caption{Relational information for all models.}
\label{fig:curved_ps_relational}
\end{figure*}

\begin{figure*}[t!]
\centering
\begin{subfigure}[b]{0.3\textwidth}
    \centering
    \includegraphics[width=\textwidth]{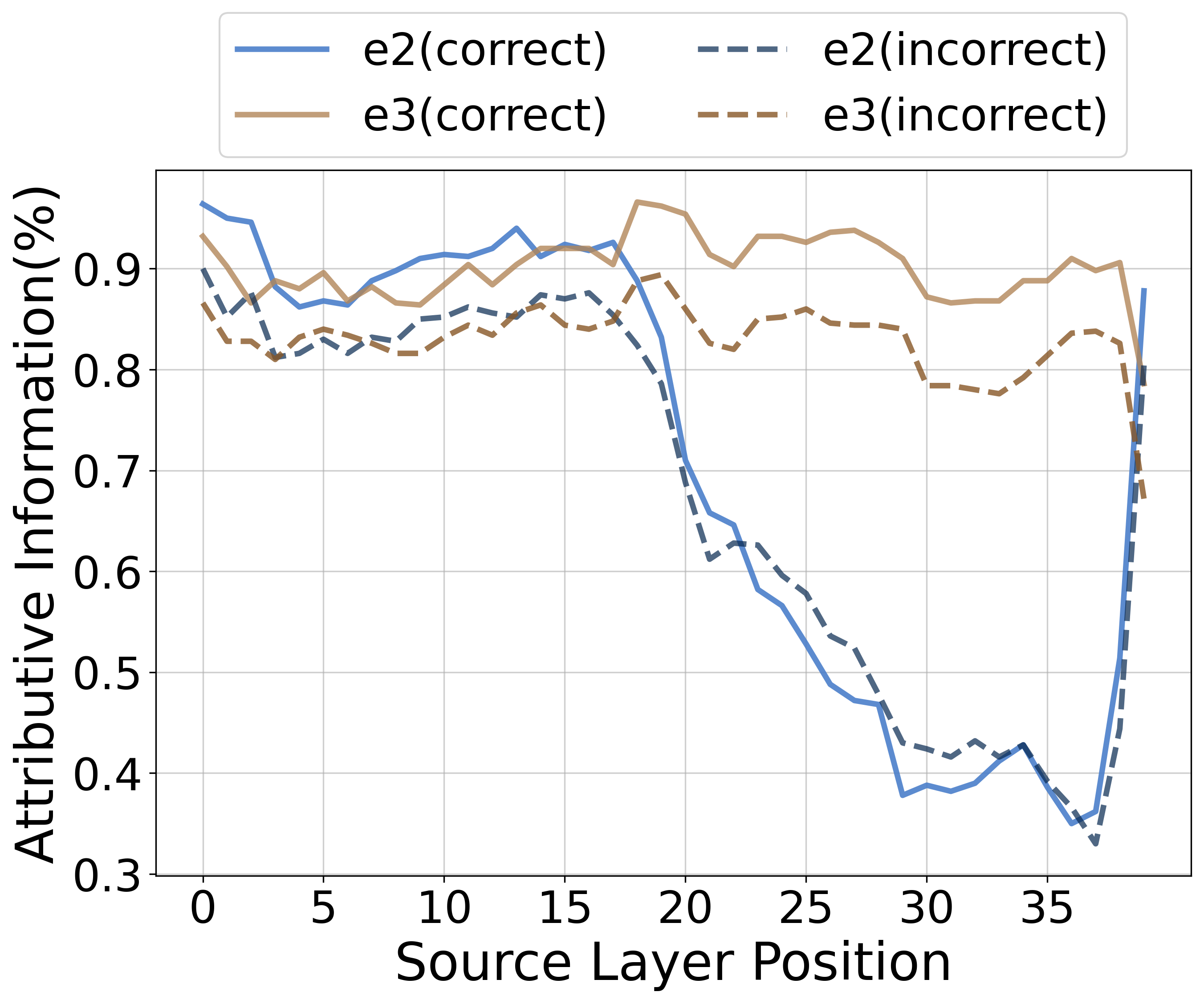}
    \caption{Llama-2-13B}
\end{subfigure}
\hfill
\begin{subfigure}[b]{0.3\textwidth}
    \centering
    \includegraphics[width=\textwidth]{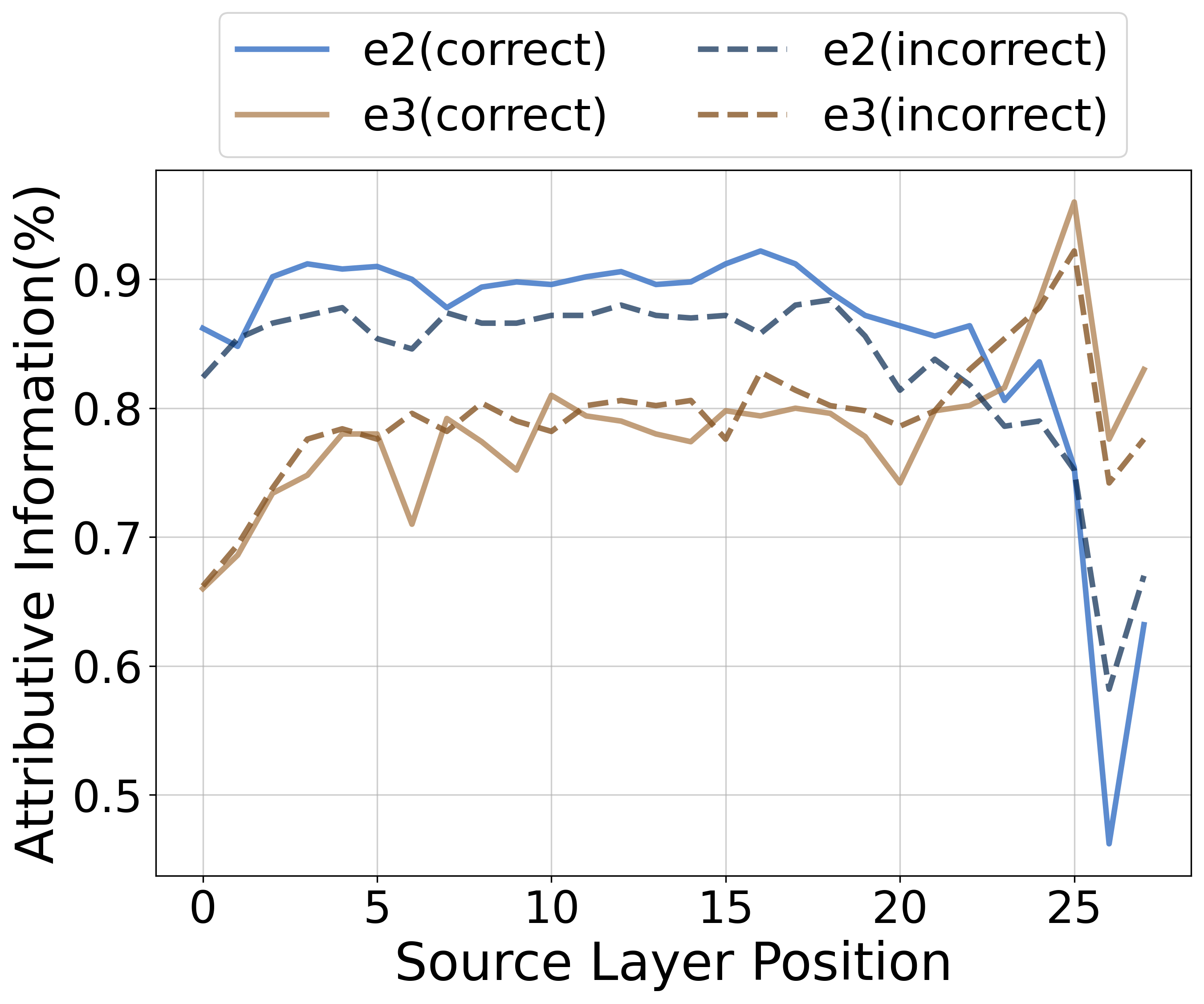}
    \caption{Gemma-7B}
\end{subfigure}
\hfill
\begin{subfigure}[b]{0.3\textwidth}
    \centering
    \includegraphics[width=\textwidth]{figures/appendix/attributive_Qwen2.5-14B.png}
    \caption{Qwen2.5-14B}
\end{subfigure}
\caption{Attributive information for all models.}
\label{fig:curved_ps_attributive}
\end{figure*}

Figure~\ref{fig:curved_ps_relational} and~\ref{fig:curved_ps_attributive} describe the proportion of cases where relational and attributive information is successfully decoded using Patchscopes for all models.
\label{appendix:curved_ps}

\section{Heatmap of Intervention Experiments}
\begin{figure*}[t!]
\centering
\begin{subfigure}[b]{0.3\textwidth}
    \centering
    \includegraphics[width=\textwidth]{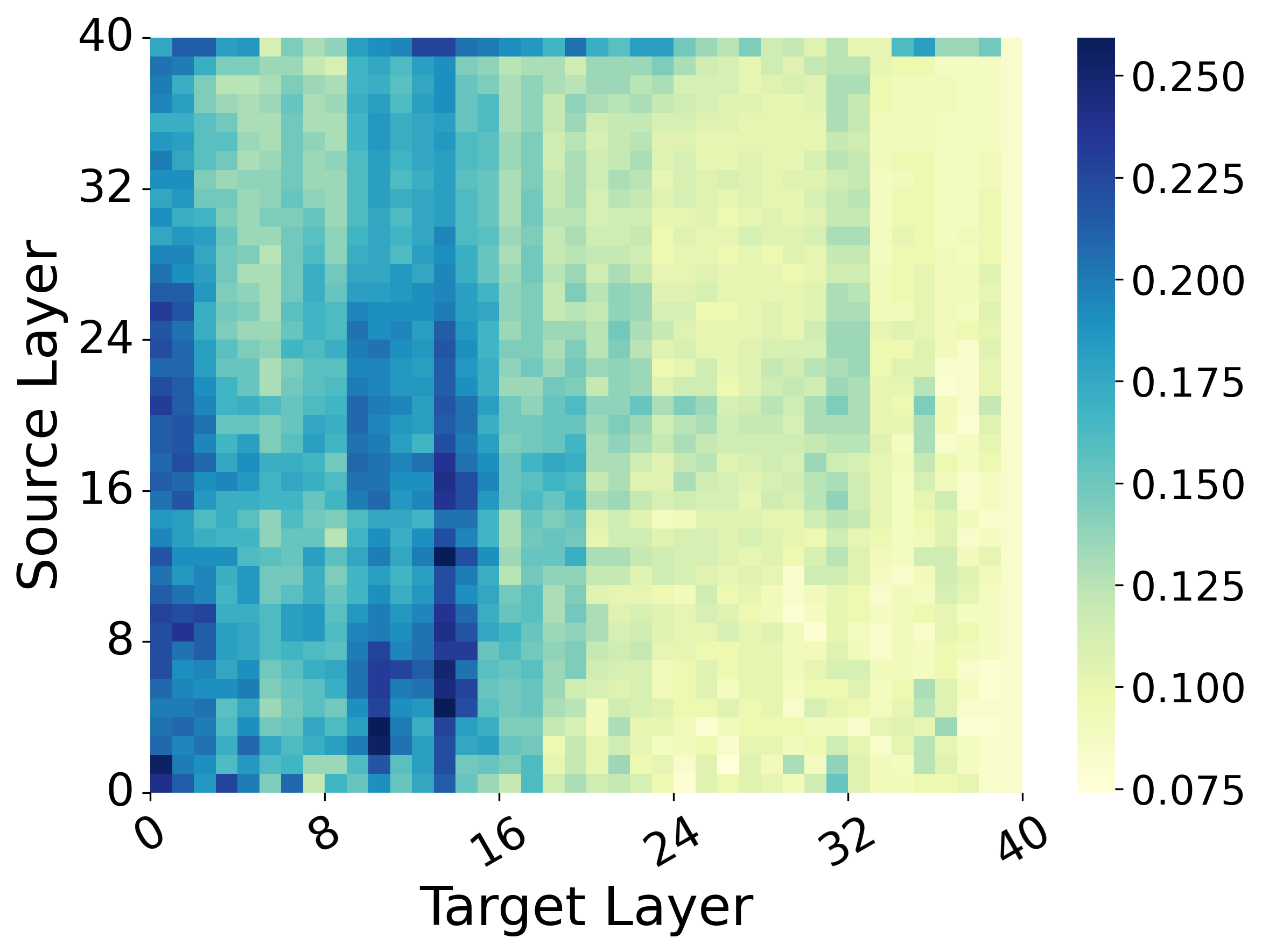}
    \caption{Llama-2-13B}
\end{subfigure}
\hfill
\begin{subfigure}[b]{0.3\textwidth}
    \centering
    \includegraphics[width=\textwidth]{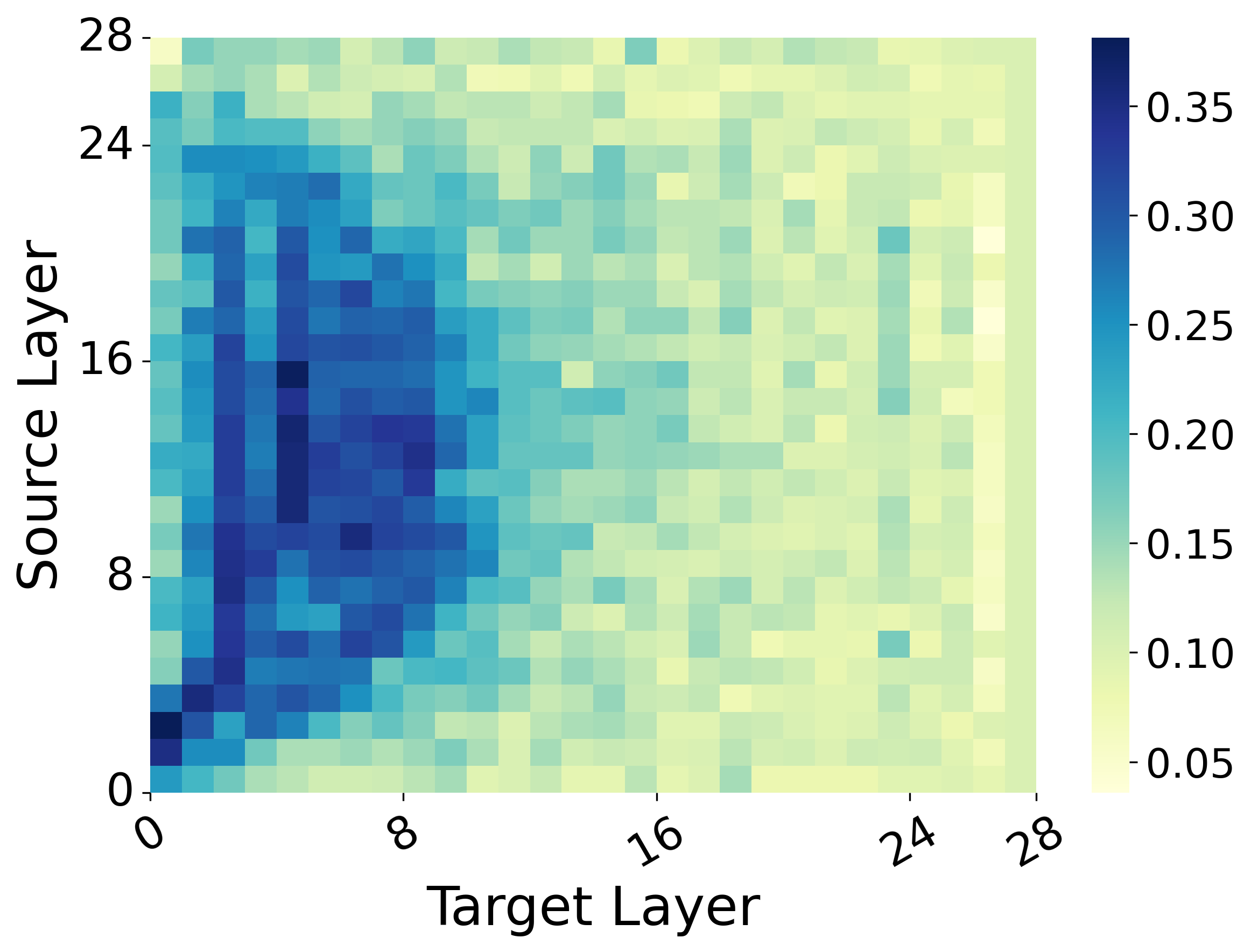}
    \caption{Gemma-7B}
\end{subfigure}
\hfill
\begin{subfigure}[b]{0.3\textwidth}
    \centering
    \includegraphics[width=\textwidth]{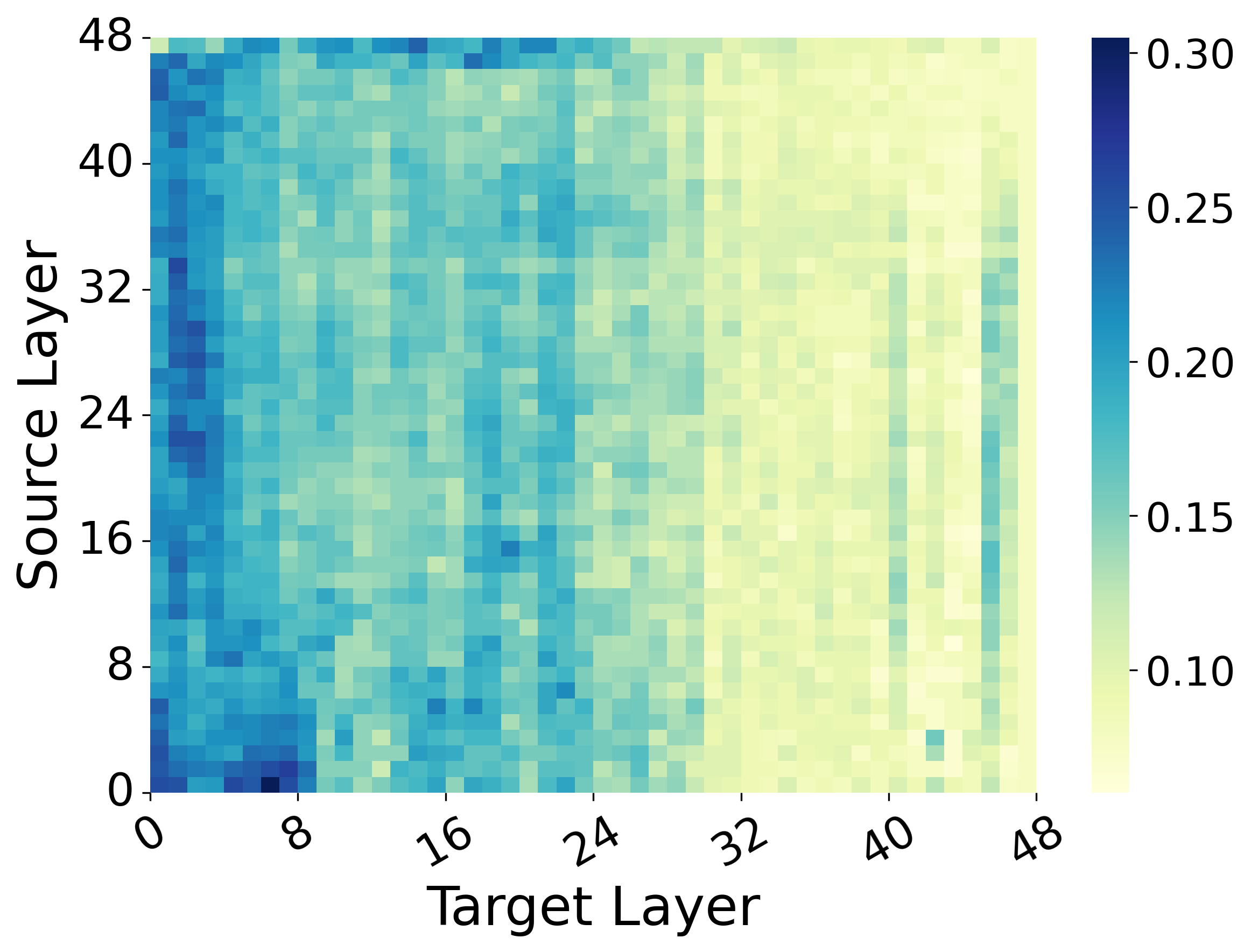}
    \caption{Qwen2.5-14B}
\end{subfigure}
\caption{Visualization of intervention experiment for all models. Source Layer refers to layers in $e_2$, and Target Layer refers to layers in the link, into which representations from $e_2$ are injected.}

\label{fig:heatmap_intervention}
\end{figure*}

Figure~\ref{fig:heatmap_intervention} visualizes head-wise contributions of the intervention experiment for all models. Patching representations to the early layers of the link proves to be particularly effective, suggesting that early contextualization of information from $e_2$ is critical for correctly applying information. 
\label{appendix:heatmap_intervention}

\section{Heatmap of Probing Experiments}
\begin{figure*}[t!]
\centering
\begin{subfigure}[b]{0.3\textwidth}
    \centering
    \includegraphics[width=\textwidth]{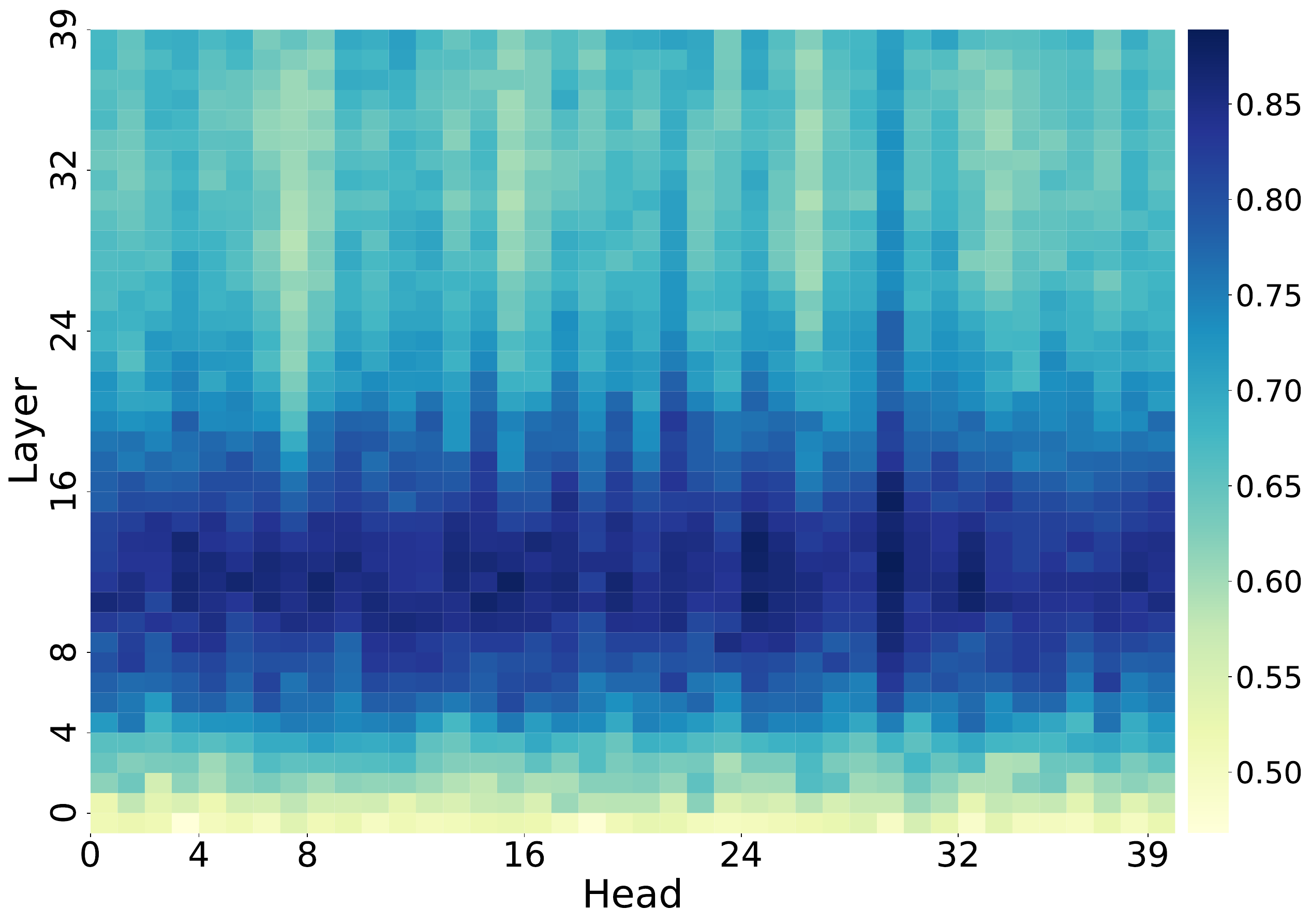}
    \caption{Llama-2-13B-chat}
\end{subfigure}
\hfill
\begin{subfigure}[b]{0.3\textwidth}
    \centering
    \includegraphics[width=\textwidth]{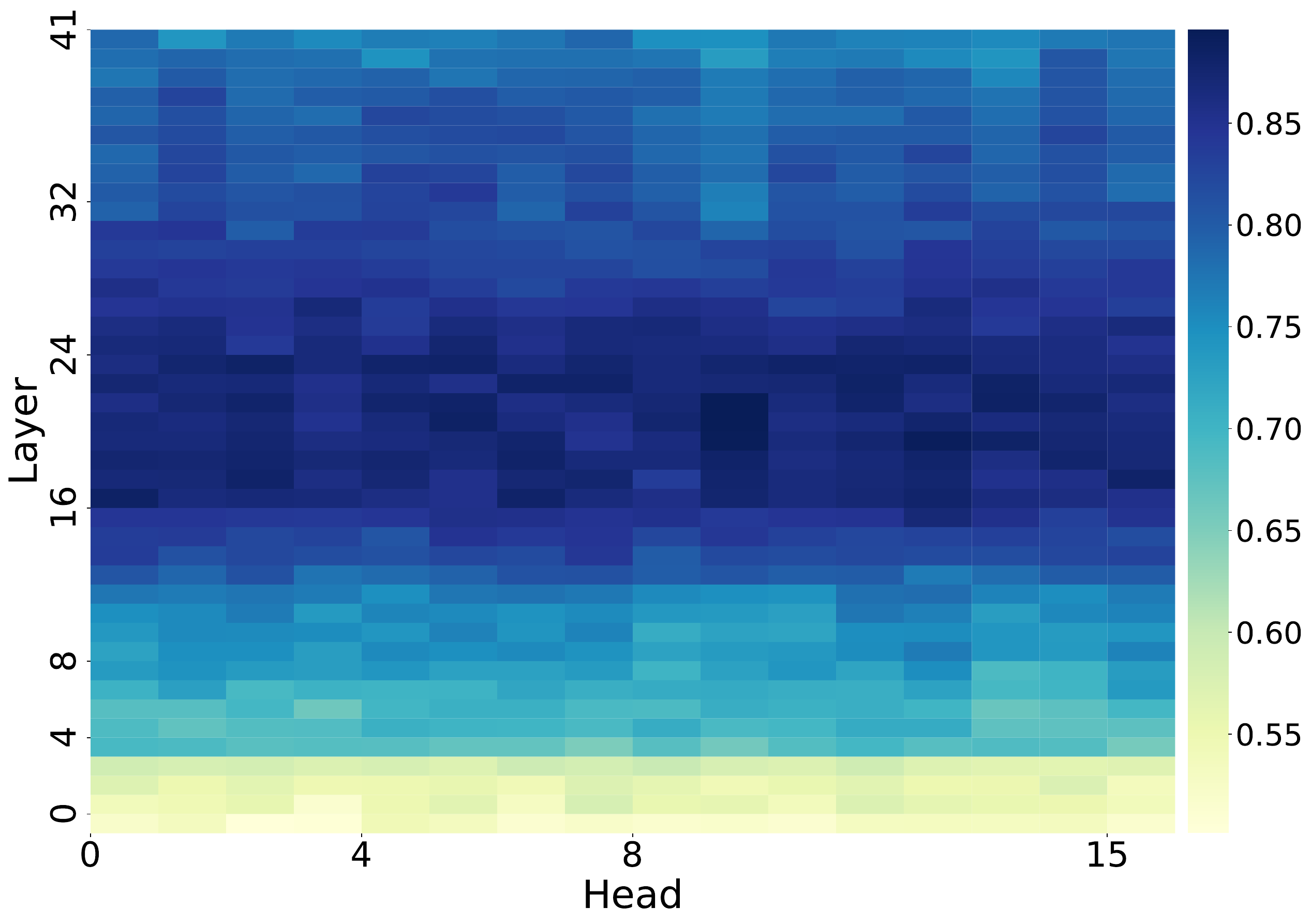}
    \caption{Gemma-2-9B-it}
\end{subfigure}
\hfill
\begin{subfigure}[b]{0.3\textwidth}
    \centering
    \includegraphics[width=\textwidth]{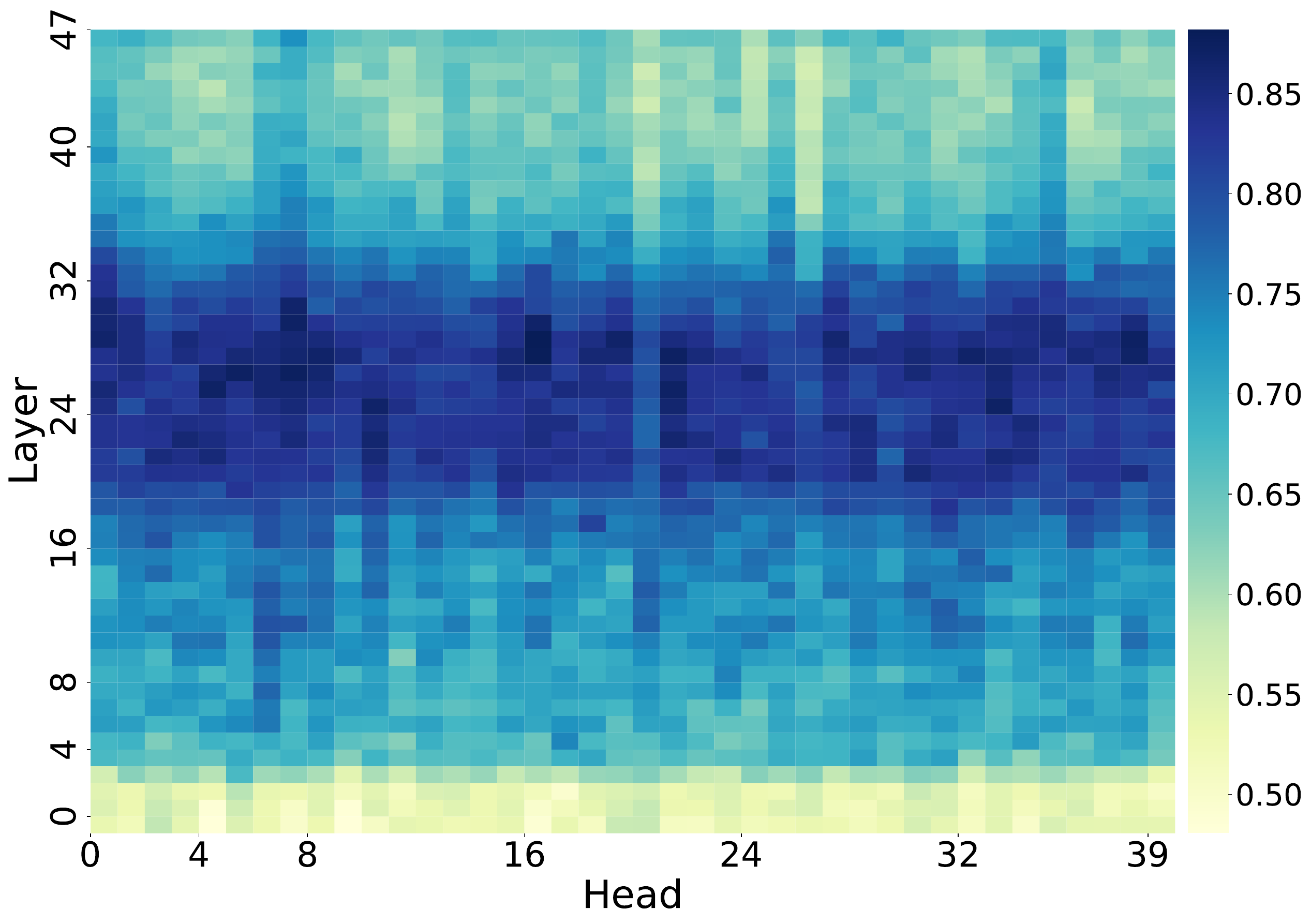}
    \caption{Qwen2.5-14B-Instruct}
\end{subfigure}
\caption{Linear probe accuracy for all models.}

\label{fig:heatmap_probing}
\end{figure*}

Figure~\ref{fig:heatmap_probing} shows heatmaps of probing experiments for all models. High accuracy in the middle layers (early-mid layers for Llama-2-13B-chat) indicates that models are capable of representing analogical structure in these regions.

\section{Relative Mutual Alignment Score}
\begin{figure*}[t!]
\centering
\begin{subfigure}[b]{0.3\textwidth}
    \centering
    \includegraphics[width=\textwidth]{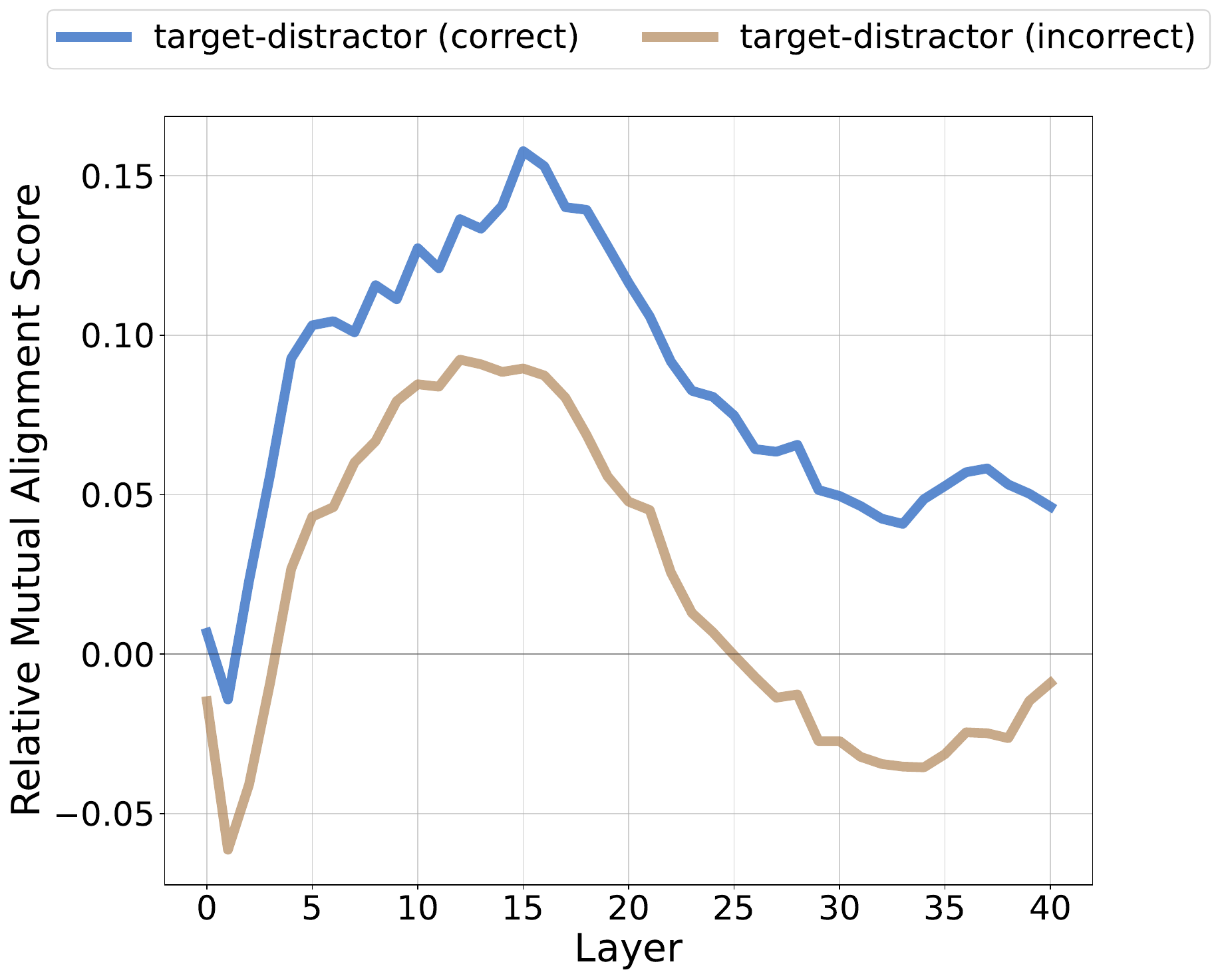}
    \caption{LLama-2-13B-chat}
\end{subfigure}
\hfill
\begin{subfigure}[b]{0.3\textwidth}
    \centering
    \includegraphics[width=\textwidth]{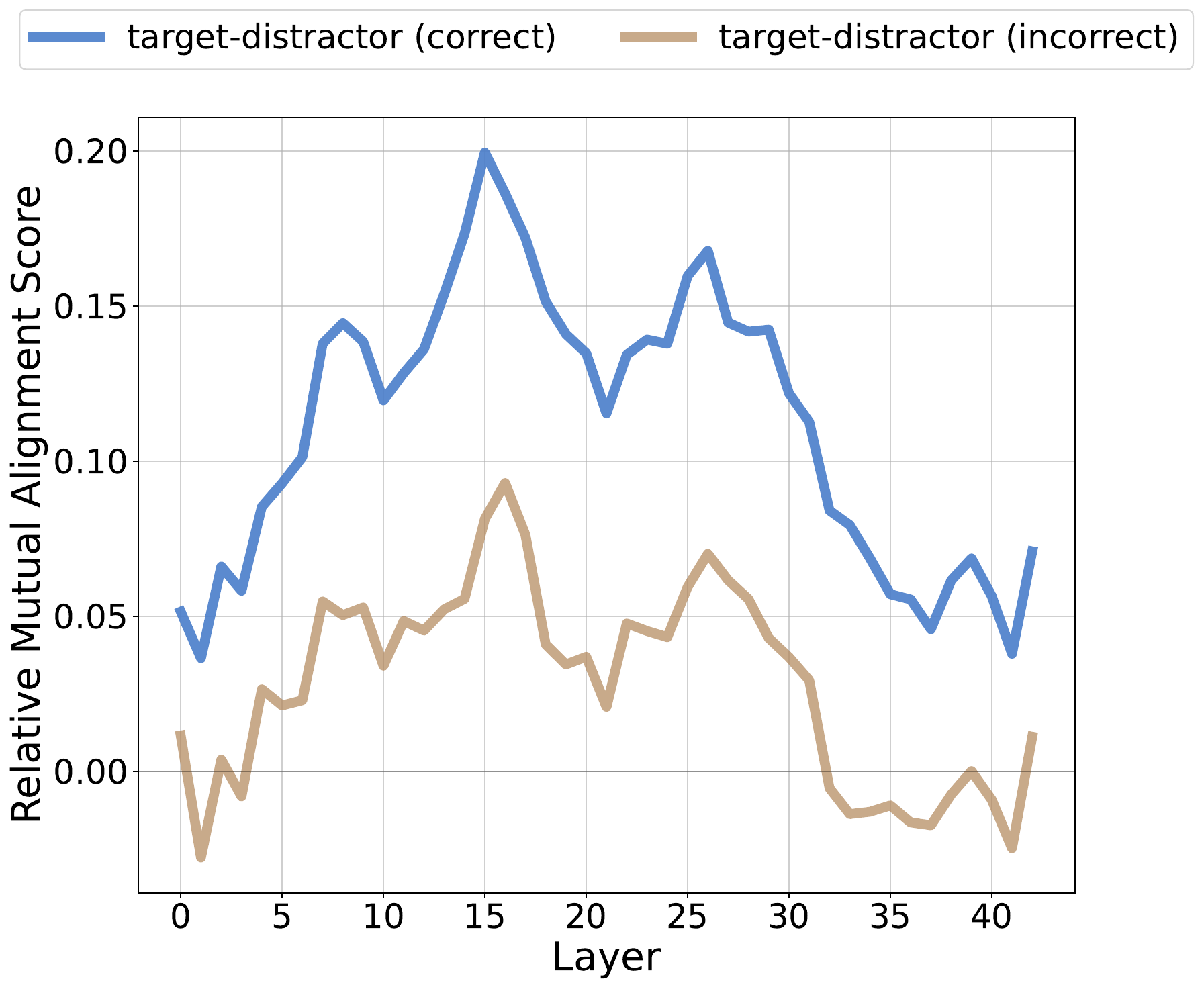}
    \caption{Gemma-2-9B-it}
\end{subfigure}
\hfill
\begin{subfigure}[b]{0.3\textwidth}
    \centering
    \includegraphics[width=\textwidth]{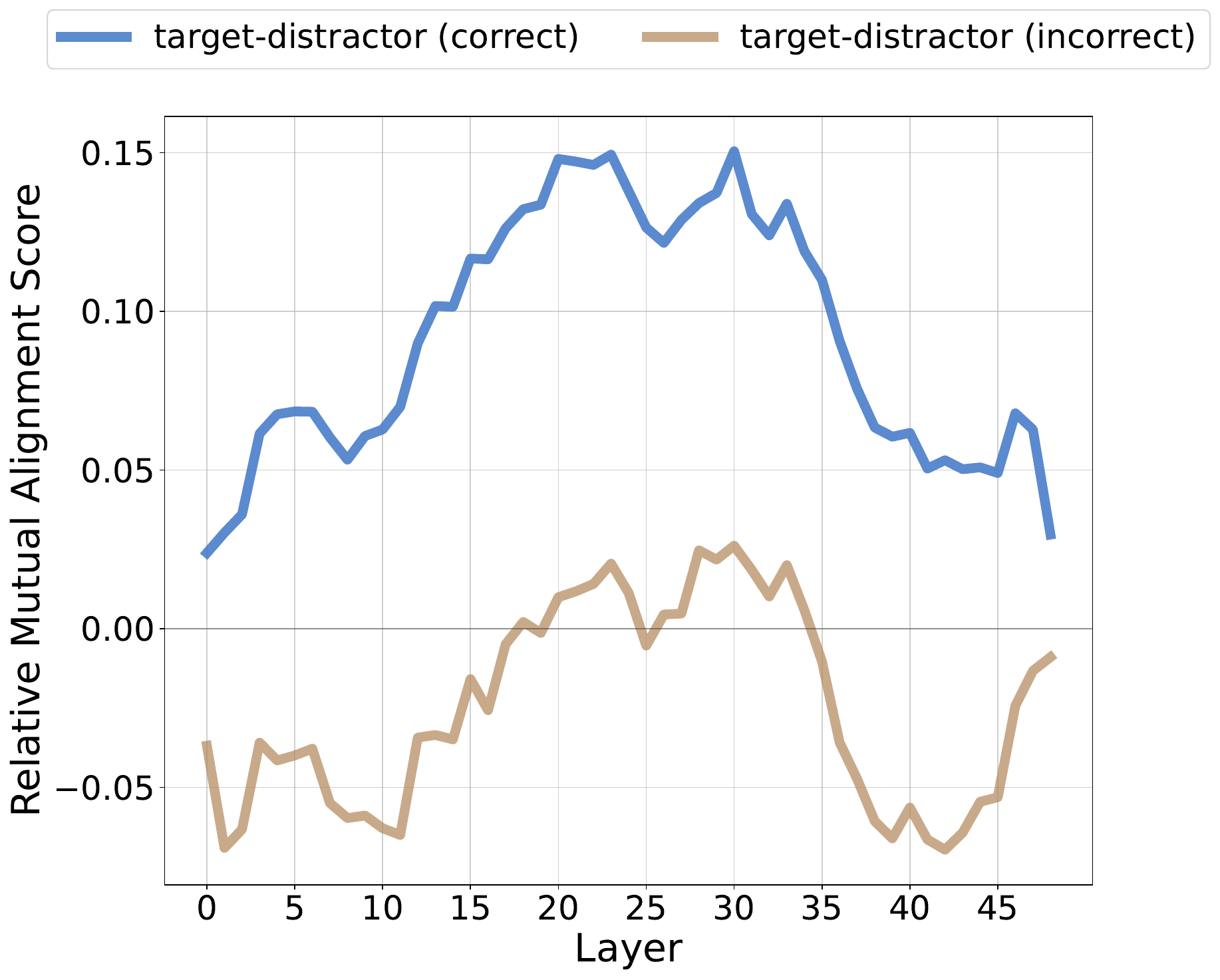}
    \caption{Qwen2.5-14B-Instruct}
\end{subfigure}
\caption{Relative mutual alignment score for all models.}

\label{fig:curved_mas}
\end{figure*}

Figure~\ref{fig:curved_mas} shows the relative mutual alignment score (MAS) for all models, computed as the difference between the MAS of source-target pairs and source-distractor pairs.
\label{appendix:curved_mas}




\begin{table}[t]
\vspace{2mm}
\centering
\footnotesize
\resizebox{\columnwidth}{!}{
\begin{tabular}{lccc}
\toprule
Model & Correct & Incorrect & Total\\
\midrule
Llama-2-13B & 809 & 924 & 1733\\
\midrule
Gemma-7B & 756 & 997 & 1753\\
\midrule
Qwen2.5-14B & 1478 & 1122 & 2600\\
\bottomrule
\end{tabular}
}
\caption{
Statistics for proportional analogy samples.
}
\label{table:dataset_proportional}
\end{table}
\begin{table}[t]
\vspace{2mm}
\centering
\footnotesize
\resizebox{\columnwidth}{!}{
\begin{tabular}{lccc}
\toprule
Model & Correct & Incorrect & Total\\
\midrule
Llama-2-13b-chat & 150 & 210 & 360\\
\midrule
Gemma-2-9B-it & 219 & 141 & 360\\
\midrule
Qwen2.5-14B-Instruct & 197 & 163 & 360\\
\bottomrule
\end{tabular}
}
\caption{
Statistics for story analogy samples.
}
\label{table:dataset_story}
\end{table}
\begin{table*}[t]
\centering
\footnotesize
\renewcommand{\arraystretch}{1.2}  
\begin{tabular}{lp{9.0cm}p{6.3cm}} 
\toprule
\textbf{Prompt Type} & \textbf{Prompt} & \textbf{Model Output} \\
\midrule

\multicolumn{3}{l}{\makecell[l]{\textit{Analogy:} “The Sign of Four is to Arthur Conan Doyle as Don Quixote is to” (answer: Miguel de Cervantes) \\ $e_1$: “The Sign of Four”, $e_2$: “Arthur Conan Doyle”, $e_3$: “Don Quixote”, $e_4$: “Miguel de Cervantes”}} \\
\cmidrule(lr){1-3}

\texttt{e2} & 
\makecell[l]{Japan is to Tokyo: capital of,\\
Theory of Evolution is to Charles Darwin: founder of,\\
Peace is to olive branch: symbol of,\\
The Sign of Four is to \textbf{x}} & 
\makecell[l]{‘, Arthur Conan: \textbf{author of}, ...’\\‘: \textbf{author of}, The Sign of Four is to ...’} \\
\cmidrule(lr){1-3}

\texttt{e3} & 
\makecell[l]{Japan is to Tokyo: capital of,\\
Theory of Evolution is to Charles Darwin: founder of,\\
Peace is to olive branch: symbol of,\\
\textbf{x} is to Miguel de Cervantes} & 
\makecell[l]{‘: \textbf{author of} Don Quixote, and so on. ...’\\‘: \textbf{author of}, x is to Don Quixote: ...’} \\
\cmidrule(lr){1-3}

\texttt{resolution} & 
\makecell[l]{Japan is to Tokyo: capital of,\\
Theory of Evolution is to Charles Darwin: founder of,\\
Peace is to olive branch: symbol of,\\
Don Quixote is \textbf{x}} & 
\makecell[l]{‘Cervantes: \textbf{author of} Don Quixote, ...’\\‘Miguel de Cervantes: \textbf{author of}, ...’} \\
\cmidrule(lr){1-3}

\makecell[l]{\texttt{resolution}\\\texttt{(default)}} & 
\makecell[l]{Syria: Country in the Middle East,\\
Leonardo DiCaprio: American actor,\\
Samsung: South Korean multinational major appliance and\\consumer electronics corporation,\\
\textbf{x}} & 
\makecell[l]{‘Miguel de Cervantes: Miguel de Cervantes was a\\Spanish writer, ...’} \\
\bottomrule
\end{tabular}
\caption{Prompts and model outputs for the relation \texttt{"author of"}.}
\label{table:author_of_prompts}
\end{table*}

\begin{table*}[t]
\centering
\footnotesize
\renewcommand{\arraystretch}{1.2}
\begin{tabular}{lp{9.0cm}p{6.3cm}}
\toprule
\textbf{Prompt Type} & \textbf{Prompt} & \textbf{Model Output} \\
\midrule

\multicolumn{3}{l}{\makecell[l]{\textit{Analogy:} “Avatar 3 is to James Cameron as Heaven \& Earth is to” (answer: Oliver Stone)\\ $e_1$: “Avatar 3”, $e_2$: “James Cameron”, $e_3$: “Heaven \& Earth”, $e_4$: “Oliver Stone”}} \\
\cmidrule(lr){1-3}

\texttt{e2} & 
\makecell[l]{Japan is to Tokyo: capital of,\\
Theory of Evolution is to Charles Darwin: founder of,\\
Peace is to olive branch: symbol of,\\
Avatar 3 is to \textbf{x}} & 
\makecell[l]{‘, James Cameron: \textbf{director of}, ...’\\‘: \textbf{director of}, Avatar 3 is to ...’} \\
\cmidrule(lr){1-3}

\texttt{e3} & 
\makecell[l]{Japan is to Tokyo: capital of,\\
Theory of Evolution is to Charles Darwin: founder of,\\
Peace is to olive branch: symbol of,\\
\textbf{x} is to Oliver Stone} & 
\makecell[l]{‘: \textbf{director of} Heaven \& Earth, and so on. ...’\\‘: \textbf{director of}, x is to Heaven \& Earth: ...’} \\
\cmidrule(lr){1-3}

\texttt{resolution} & 
\makecell[l]{Japan is to Tokyo: capital of,\\
Theory of Evolution is to Charles Darwin: founder of,\\
Peace is to olive branch: symbol of,\\
Heaven \& Earth is \textbf{x}} & 
\makecell[l]{‘Oliver Stone: \textbf{director of},’\\‘Oliver Stone is to JFK: \textbf{director of}, ...’} \\
\cmidrule(lr){1-3}

\makecell[l]{\texttt{resolution}\\\texttt{(default)}} &
\makecell[l]{Syria: Country in the Middle East,\\
Leonardo DiCaprio: American actor,\\
Samsung: South Korean multinational major appliance and\\consumer electronics corporation,\\
\textbf{x}} & 
\makecell[l]{‘Oliver Stone: Oliver Stone is an American \\filmmaker,director of many political films, ...’} \\
\bottomrule
\end{tabular}
\caption{Prompts and model outputs for the relation \texttt{"director of"}.}
\label{table:director_of_prompts}
\end{table*}

\begin{table*}[t]
\centering
\footnotesize
\renewcommand{\arraystretch}{1.2}
\begin{tabular}{lp{9.0cm}p{6.3cm}}
\toprule
\textbf{Prompt Type} & \textbf{Prompt} & \textbf{Model Output} \\
\midrule

\multicolumn{3}{l}{\makecell[l]{\textit{Analogy:} “True Romance is to Hans Zimmer as Fugue in G minor, ‘Little’, BWV 578 is to” (answer: Johann Sebastian Bach)\\ $e_1$: “True Romance”, $e_2$: “Hans Zimmer”, $e_3$: “Fugue in G minor, ‘Little’, BWV 578”, $e_4$: “Johann Sebastian Bach”}} \\
\cmidrule(lr){1-3}

\texttt{e2} & 
\makecell[l]{Japan is to Tokyo: capital of,\\
Theory of Evolution is to Charles Darwin: founder of,\\
Peace is to olive branch: symbol of,\\
True Romance is to \textbf{x}} & 
\makecell[l]{‘, Hans Zimmer: \textbf{composer of}, ...’\\‘: \textbf{composer of}, True Romance is to ...’} \\
\cmidrule(lr){1-3}

\texttt{e3} & 
\makecell[l]{Japan is to Tokyo: capital of,\\
Theory of Evolution is to Charles Darwin: founder of,\\
Peace is to olive branch: symbol of,\\
\textbf{x} is to Johann Sebastian Bach} & 
\makecell[l]{‘: \textbf{composer of} Fugue in G minor, and so on. ...’\\‘: \textbf{composer of}, x is to Fugue in G minor: ...’} \\
\cmidrule(lr){1-3}

\texttt{resolution} & 
\makecell[l]{Japan is to Tokyo: capital of,\\
Theory of Evolution is to Charles Darwin: founder of,\\
Peace is to olive branch: symbol of,\\
Fugue in G minor, ‘Little’, BWV 578 is \textbf{x}} & 
\makecell[l]{‘Johann Sebastian Bach: \textbf{composer of}, ...’\\‘J. S. Bach: \textbf{composer of}, ...’} \\
\cmidrule(lr){1-3}

\makecell[l]{\texttt{resolution}\\\texttt{(default)}} & 
\makecell[l]{Syria: Country in the Middle East,\\
Leonardo DiCaprio: American actor,\\
Samsung: South Korean multinational major appliance and\\consumer electronics corporation,\\
\textbf{x}} & 
\makecell[l]{‘Johann Sebastian Bach: Bach was a German \\composer and musician of the Baroque period, ...’} \\
\bottomrule
\end{tabular}
\caption{Prompts and model outputs for the relation \texttt{"composer of"}.}
\label{table:composer_of_prompts}
\end{table*}


\end{document}